\documentclass{article}

    \usepackage[preprint]{neurips_2026}

\usepackage[utf8]{inputenc} 
\usepackage[T1]{fontenc}    
\usepackage{hyperref}       
\usepackage{url}            
\usepackage{booktabs}       
\usepackage{amsfonts}       
\usepackage{nicefrac}       
\usepackage{microtype}      
\usepackage{xcolor}         
\usepackage{listings}
\usepackage{amsmath,amssymb,amsfonts}
\usepackage{amsthm}
\usepackage{enumitem}
\usepackage{bbm}
\usepackage{algorithm}
\usepackage{algpseudocode}
\usepackage{xspace}
\usepackage{graphicx}
\usepackage{thmtools}
\usepackage{thm-restate}
\usepackage{compact_}
\usepackage{wrapfig}
\usepackage{titletoc}   
\usepackage{etoolbox}   
\usepackage{multirow}
\usepackage{float}
\usepackage{subcaption}
\usepackage{verbatim}

\usepackage{threeparttable}

\definecolor{tagcolor}{rgb}{0,0,0}
\definecolor{attrcolor}{rgb}{0.6,0,0.8}
\definecolor{valcolor}{rgb}{0,0,1}

\lstdefinelanguage{CustomXML}{
  morekeywords={/>},
  morecomment=[s]{<!--}{-->},
  morestring=[b]",
  sensitive=true,
}

\theoremstyle{plain}

\theoremstyle{definition}

\theoremstyle{remark}

\usepackage[utf8]{inputenc} 
\usepackage[T1]{fontenc}    
\usepackage{hyperref}       
\usepackage{url}            
\usepackage{booktabs}       
\usepackage{amsfonts}       
\usepackage{nicefrac}       
\usepackage{microtype}      
\usepackage{xcolor}         

\usepackage{array}
\usepackage{tcolorbox}
\tcbuselibrary{skins, breakable, listings, theorems}
\usepackage{caption}
\usepackage{adjustbox} 
\usepackage{float} 
\usepackage{soul}
\usepackage{makecell}
\usepackage{siunitx}
\usepackage{colortbl}
\usepackage{multicol}
\usepackage{multirow}
\usepackage{arydshln}
\usepackage{graphicx}
\usepackage{subcaption}
\usepackage{amsmath}
\usepackage{wrapfig}
\tcbuselibrary{skins, breakable}
\usepackage{enumitem}
\usepackage{xcolor}
\usepackage[dvipsnames]{xcolor} 

\definecolor{ThemeGreen}{RGB}{115, 140, 50}

\newcommand{\sectiontitle}[1]{%
    \vspace{1ex}%
    \noindent\textcolor{ThemeGreen}{\large\bfseries #1}\par\vspace{0.5ex}%
}

\newcommand{\innerline}{%
    \vspace{1ex}\noindent\textcolor{gray!50}{\rule{\linewidth}{0.5pt}}\vspace{1ex}%
}
\newtcolorbox{customcase}[1][]{
    colback=white,             
    colframe=ThemeGreen,       
    colbacktitle=ThemeGreen,   
    coltitle=white,            
    fonttitle=\bfseries\large, 
    title=#1,                  
    arc=2mm,                   
    boxrule=1.5pt,             
    left=3mm, right=3mm,       
    top=2mm, bottom=2mm,       
    breakable                  
}
\title{NaiAD: Initiate Data-Driven Research\\ for LLM Advertising}

\author{%
   Yihang Zhang \thanks{This work was done by the author during his internship at Tsinghua University.} \\
  Tsinghua University\\
  Beijing, China \\
  \texttt{hemmaxacand@gmail.com} \\
  \And
  Zimeng Huang \\
  College of AI \\
  Tsinghua University \\
  Beijing, China\\
  \texttt{simona22336098@gmail.com} \\
  \And 
  Ren Zhai \\
  Department of Literature, \\Arts and Communication \\
  Anhui International Studies University \\
  Anhui, China \\
  \texttt{zren@stu.aisu.edu.cn}
  \And
  Yipeng Kang \thanks{Correspondence to: Tonghan Wang <twang1@g.harvard.edu>, Yipeng Kang <kangyipeng@bigai.ai>.} \\
  State Key Laboratory of General \\
  Artificial Intelligence, BIGAI \\
  Beijing, China \\
  \texttt{kangyipeng@bigai.ai}
  \And
  Tonghan Wang $^\dagger$ \\
  College of AI \\
  Tsinghua University \\
  Beijing, China \\
  \texttt{twang1@g.harvard.edu}
}

\begin{document}

\maketitle

\begin{abstract}

Reconciling platform revenue with user experience in LLM advertising motivates a data-centric foundation.
We introduce \textbf{NaiAD}, the first comprehensive dataset for LLM-native advertising comprising $58,999$ carefully constructed ad-embedded responses paired with user queries. NaiAD is organized around theoretically grounded evaluation metrics that separately and comprehensively capture user and commercial utility. To mitigate the dimensional collinearity of aligned LLMs, we propose a decoupled generation pipeline that produces structurally diverse samples, ranging from responses that explicitly disentangle stakeholder utilities to responses that are uniformly strong or weak across dimensions. We further provide score labels calibrated by a Variance-Calibrated Prediction-Powered Inference (VC-PPI) framework, aligning automated scoring with human annotations. Mechanistic analyses reveal that successful ad integration relies on reasoning paths that cluster into four distinct semantic strategies. Models leveraging NaiAD internalize these strategies to simultaneously improve user and commercial utility, while enabling independent control over these distinct objectives via in-context learning. Together, these results position NaiAD as a foundational infrastructure for developing future LLM-native ad systems. \\
{\centering \textbf{NaiAD collection: \url{https://huggingface.co/datasets/MaxAcand/NaiAD}}}

\end{abstract}

\section{Introduction}
\label{sec:intro}

Recently, the integration of advertising into Large Language Model (LLM) responses by prominent AI organizations has marked a pivotal shift in the generative AI ecosystem \cite{openai_ad}. While approached with careful design and sophisticated fallback mechanisms, this initiative has encountered measurable user hesitation \cite{observer2026openaiads,ndtv2026backlash}. This friction underscores a fundamental tension: balancing the imperative to monetize capital-intensive models with the user's desire for uninterrupted conversational experiences. 

Advertising is not inherently incompatible with LLM-based interaction. Generative models can enable advertising that is clearly labeled, context-aware, helpful, and unobtrusive. Yet, current scalable systems frequently treat ads as external inserts, mechanically appending banners or sponsored links to generated responses \cite{xu2026ad,hu2025gem,openai_ad,balseiro2026position}. This semantic disconnect disrupts conversational coherence, harms user experience, and ultimately reduces advertising effectiveness.

This inefficiency reflects a gap in the existing research. Prior work largely focuses on the economic and algorithmic aspects of LLM advertising, such as adapting search auction mechanisms to generative settings \cite{chatgpt_replace_search,llm_survey,shi2025know} and optimizing bidding and pricing \cite{feizi2023online, chen_budget-constrained_2024,duetting2024mechanism,balseiro2026position}. Consequently, the generative quality of the sponsored content itself remains underexplored. However, user satisfaction and monetization are not mutually exclusive \cite{zhao2025llm}. High-quality native sponsorships can align both objectives, akin to how skilled human creators seamlessly integrate ads into their content.



These observations point to a central bottleneck in generative advertising: the absence of a data-centric foundation for studying, evaluating, and training high-quality conversational ads. Effective LLM advertising requires models to learn how different integration strategies affect user utility, perceived naturalness, and platform revenue. Such learning necessitates a dataset satisfying two key desiderata. 
(1) \textbf{Multi-dimensional, unbiased assessment.} Sponsored responses should be evaluated across decoupled axes reflecting multiple stakeholders: \textit{User Utility}, measuring whether the user's query is answered naturally and coherently; and \textit{Commercial Utility} for advertisers and platforms, assessing if brand information is effectively integrated and likely to drive user engagement.
(2) \textbf{Structural diversity and hard negatives.} The dataset should contain diverse and disentangled samples. For instance, high user utility does not automatically imply high commercial value. The dataset should therefore include controlled ``hard negatives'' spanning different combinations of user and commercial quality, preventing both generators and evaluators from exploiting spurious correlations.

\begin{figure*}[t]
    \centering
    \includegraphics[width=\textwidth]{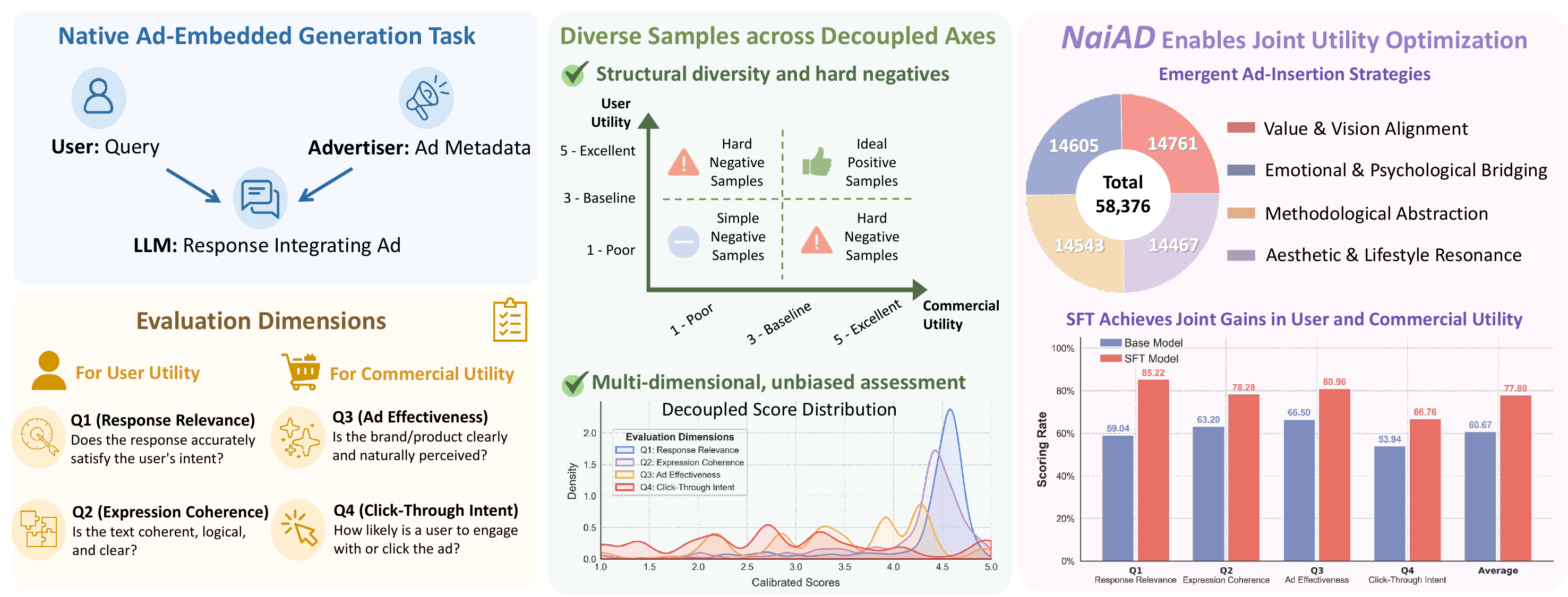} 
    \caption{\textbf{An overview of our NaiAD dataset.} 
    \textbf{(Left)} We define the task and the four decoupled evaluation dimensions. 
    \textbf{(Middle)} We illustrate our data-centric methodology, emphasizing the generation of structurally diverse samples including ``hard negatives'' to break dimensional collinearity. The resulting decoupled score distributions confirm the successful creation of a dimensionally-orthogonal dataset.
    \textbf{(Right)} We present our main findings: (Top) The discovery that LLM ad-insertion behavior converges into four emergent strategies. (Bottom) Empirical validation showing that Supervised Fine-Tuning (SFT) on NaiAD enables a base model to achieve significant joint gains across all utility dimensions, proving the dataset's effectiveness.}
    \label{fig:overview}
\end{figure*}

We introduce \textbf{NaiAD} (Native Ad Integration and Assessment Dataset), the first comprehensive dataset for LLM-based native advertising. NaiAD is designed around the two desiderata and comprises $58,999$ carefully constructed ad-embedded responses: $58,376$ LLM-generated responses and $623$ YouTube-sourced human responses. Grounded in Jakobson's semiotic communication theory \citep{Jakobson1960Closing} and Austin and Searle's speech act theory \citep{austin_how_1975,searle_speech_2012}, we evaluate these responses across four decoupled metrics: \textit{Response Relevance} and \textit{Expression Coherence} (measuring User Utility), alongside \textit{Ad Effectiveness} and \textit{Click-Through Intent} (measuring Commercial Utility). However, dataset construction in this setting is challenging because of dimensional collinearity: aligned LLMs tend to produce outputs that are uniformly strong or weak across evaluation metrics. To mitigate this, we design a multi-dimensional decoupled generation pipeline to produce systematically diverse ``hard negatives''. Furthermore, to evaluate at scale without prohibitive annotation costs or evaluator bias, we introduce a Variance-Calibrated Prediction-Powered Inference (VC-PPI) framework, utilizing a human-annotated subset to statistically align LLM-based evaluations with human judgment.

Using NaiAD, we can reconcile user and commercial utility. Pareto optimality analysis reveals that samples produced by the multi-dimensional decoupled generation pipeline  consistently outperform YouTube-sourced human samples in balancing these objectives. Moreover, Supervised Fine-Tuning (SFT) on high-quality NaiAD subsets enables base models to simultaneously improve user utility and Click-Through Rate (CTR). Furthermore, In-Context Learning (ICL) experiments demonstrate that the dataset's structural diversity empowers LLMs with decoupled controllable generation, enabling them to independently adjust user and commercial utility to meet specific multi-dimensional target profiles. Mechanistically, we uncover that successful native ad integration relies on constructing a ``Logical Bridge'': an internal reasoning path linking the user's query to the advertisers' core value. Analyzing thousands of such bridges reveals an emergent low-dimensional structure where ad integration strategies converge into four semantic clusters. 

Figure~\ref{fig:overview} summarizes our dataset and key findings. In summary, our main contributions are three-fold:
\begin{itemize}[leftmargin=*, topsep=0pt, itemsep=2pt]
    \item We release \textbf{NaiAD}, the first dimensionally-orthogonal dataset for LLM ad-embedded generation, constructed via a decoupled generation pipeline and evaluated using our VC-PPI calibration framework to eliminate dimensional collinearity and evaluator bias.
    \item Our empirical analysis reveals four successful generative strategies behind native LLM advertising: embeddings of the ``Logical Bridges'' used to harmonize user and commercial utility converge into four distinct clusters in a low-dimensional semantic space.
    \item We demonstrate through Pareto optimality analysis, SFT, and In-Context Learning that user and commercial utility can be both jointly optimized and independently controlled, establishing a robust foundation for future generative advertising systems.
\end{itemize}




\section{Empirical Insight: The Logical Bridge and Strategy Convergence}
\label{sec:empirical_insight}

\begin{figure*}[t]
    \centering
    \includegraphics[width=\textwidth]{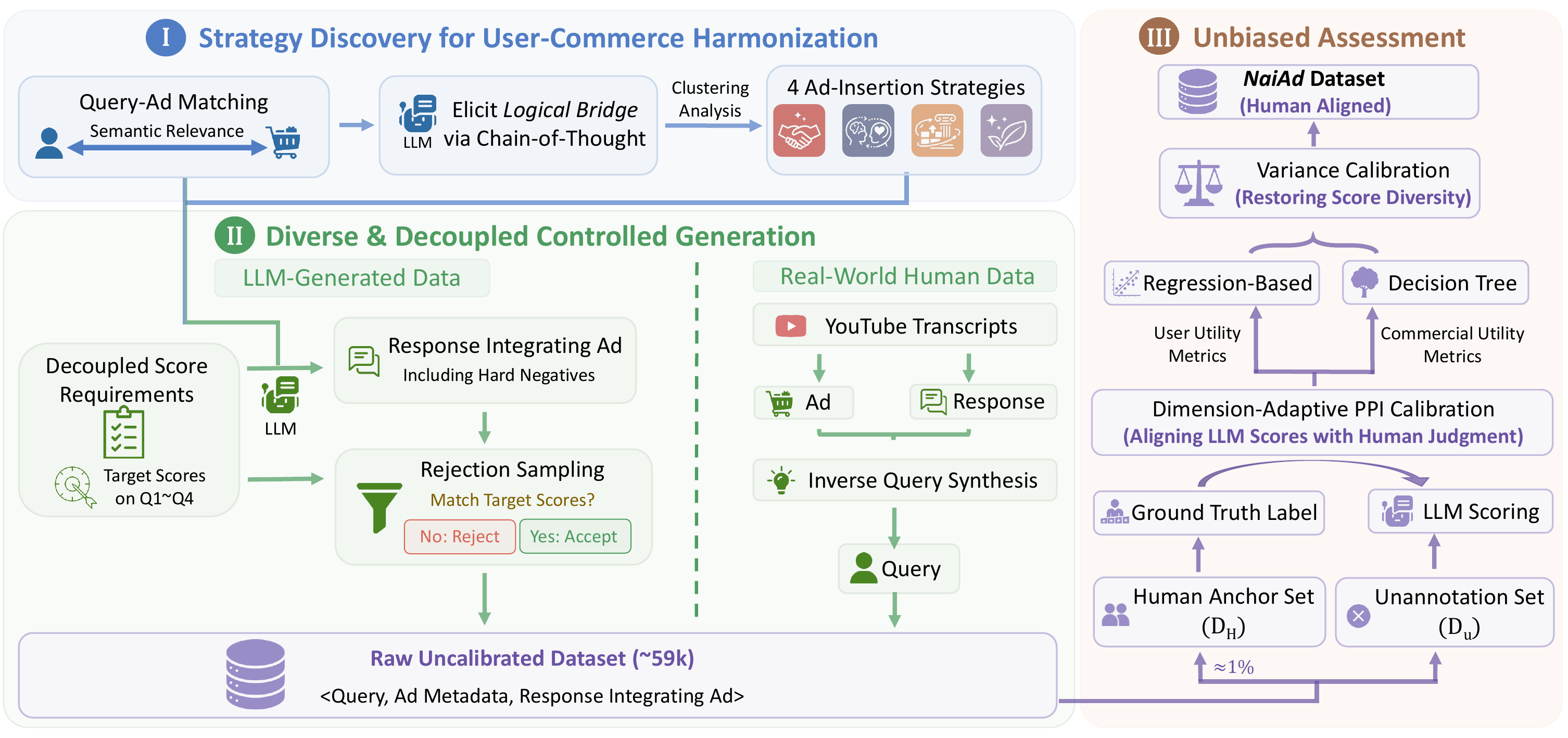} 
    \caption{\textbf{The construction and calibration pipeline of NaiAD.} \textbf{Phase I:} Eliciting and clustering LLM reasoning paths to discover four core ad-insertion strategies. \textbf{Phase II:} A decoupled generation phase creating structurally diverse raw data via target-constrained rejection sampling for LLMs, running parallel to inverse query synthesis for human transcripts. \textbf{Phase III:} A Dimension-Adaptive Prediction-Powered Inference (PPI) and variance calibration framework that aligns raw LLM evaluations with human judgments to produce the final unbiased dataset.}
    \label{fig:pipeline}
\end{figure*}

To construct the NaiAD dataset proposed in Section \ref{sec:intro}, our primary objective is to generate responses that effectively balance user utility and commercial utility. However, a major challenge immediately arises: mechanically forcing an ad into a conversation inevitably causes user frustration. Before generating the dataset at scale, we must first decipher the fundamental generative mechanism of \textit{how} to produce high-quality, harmonized samples. 

We hypothesize that successfully breaking the trade-off between user and commercial utility requires the model to construct a \textbf{``Logical Bridge''}---a latent reasoning path identifying a natural semantic intersection between the user's query and the advertiser's value proposition. To harness this, we prompt the LLM on a small batch of data to explicitly articulate these reasoning paths and perform structural analysis to uncover the emergent, heuristic strategies behind native ad integration. In this section, we deconstruct how LLMs cognitively perform this harmonization, using these discovered strategies to guide the generation process of our subsequent dataset.

\subsection{Query-Ad Matching and Logical Bridge Construction}

To investigate how models construct these bridges, we first establish a foundation of semantic relevance by pairing $1,986$ real-world queries from INFINITY-CHAT \citep{jiang2025artificial} with the most relevant advertisements from the AVTI\footnote{\url{https://github.com/Agentyzu/MAE-AM}} pool. Pairs are determined using \texttt{paraphrase-multilingual-MiniLM-L12-v2} embeddings \citep{reimers-2019-sentence-bert} and cosine similarity (see Appendix \ref{sec:data_sources} for details).

Once we have these contextually grounded pairs, we seek to understand \textit{how} the LLM seamlessly connects them. Using a highly capable generative model (detailed in Appendix \ref{app:model_details}), we prompt it to explicitly articulate the Logical Bridge. By forcing the model to plan the transition from user intent to brand value, we transform a hidden process into observable text, ensuring the ad integration is strategically grounded and transparent.

\subsection{Discovering the Four Core Ad-Insertion Strategies}

\begin{figure}[htbp]
    \centering
    \includegraphics[width=\textwidth]{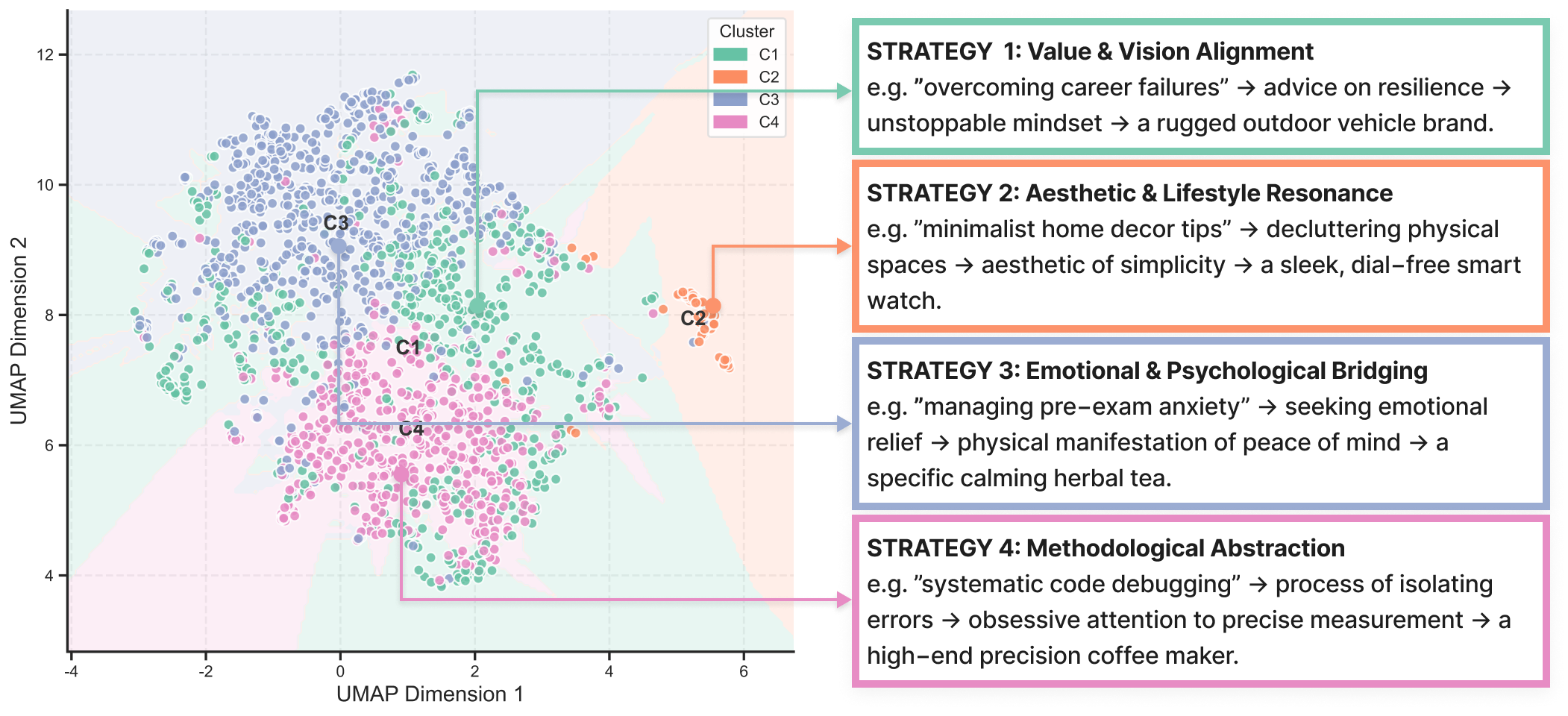}
    \caption{\textbf{Latent space visualization of Logical Bridges via UMAP.} The 30D space reveals non-overlapping regions corresponding to the four elicited strategies, illustrating how the LLM shifts its reasoning to natively integrate ads. Because constructing a Logical Bridge inherently requires complex conceptual association, we adopt the structured framework of Association Reasoning Paths \citep{huang2025mm} to represent these transitions as directed multi-hop sequences.}
    \label{fig:couple}
\end{figure}

To determine if LLM ad-embedding follows predictable patterns, we convert the generated textual bridges into Sentence-BERT embeddings. Because distance-based clustering algorithms often struggle with high-dimensional text embeddings, we use Principal Component Analysis (PCA) to progressively reduce the vectors into a condensed 30-dimensional space, removing noise while preserving semantic groupings. Subsequent K-Means clustering in this space reveals an optimal structure at $K=4$. As detailed in Appendix \ref{app:clustering}, this determination is substantiated by a clear elbow in the Sum of Squared Errors (SSE) curve and, more definitively, by a global peak in the Silhouette Score, which provides a rigorous metric for cluster separation and cohesion. This convergence suggests that \textbf{LLM ad-insertion behavior converges into four distinct cognitive strategies.} Projecting these embeddings into a 2D space via UMAP (Figure \ref{fig:couple}) allows us to categorize these strategies.

\begin{itemize}[leftmargin=*, topsep=0pt, itemsep=2pt]
   \item \textbf{Strategy 1: Value \& Vision Alignment (The ``Mindset'' Bridge).} The LLM elevates the query to a macroscopic philosophy, connecting it to an advertiser that shares identical systemic values.
    
    \item \textbf{Strategy 2: Aesthetic \& Lifestyle Resonance (The ``Vibe'' Bridge).} The LLM bypasses literal topics to match the stylistic or aesthetic constraints of the query with a brand sharing an identical lifestyle identity.
    
    \item \textbf{Strategy 3: Emotional \& Psychological Bridging (The ``Empathy'' Bridge).} The LLM identifies the underlying emotional driver and positions the product as a tangible means of emotional relief or enhancement.
    
    \item \textbf{Strategy 4: Methodological Abstraction (The ``Craftsmanship'' Bridge).} Through cross-domain feature migration, the model isolates the operational rigor required by the user's task and links it to a product characterized by similar precision or craftsmanship.
\end{itemize}

We providd {case studies} on these strategies in Appendix~\ref{app:case_studies}. Our empirical analysis demonstrates that when projected into a low-dimensional space, embeddings of ``Logical Bridges'' naturally organize into four distinct clusters, remarkably revealing four successful generative strategies underpinning native LLM advertising.
\section{The NaiAD Dataset: Multi-Dimensional Decoupled Generation}
\label{sec:controlled_generation}

Having demystified the underlying generation strategies, we proceed to the core construction of the NaiAD dataset. The end-to-end pipeline for discovering integration strategies, generating diverse samples, and conducting unbiased assessment is illustrated in Figure~\ref{fig:pipeline}. 

\textbf{Task Formalization.} We define the \textit{Native Ad-Embedded Generation} task as follows: given a user query $Q$ and advertisement metadata $A$, a model must generate a response $R$ that fulfills $Q$ while seamlessly integrating $A$ without breaking conversational coherence.

Constructing a comprehensive dataset for this task requires ensuring both \textbf{high-quality semantic integration} and \textbf{structural diversity}. Relying on unguided generation often yields generic, low-quality insertions. Furthermore, aligned models suffer from \textbf{Dimensional Collinearity (The Halo Effect)} \citep{jiang2025artificial}, producing outputs that are uniformly high or low across all metrics, thereby lacking the ``hard negatives'' needed to train discriminative boundaries. To address these challenges, we utilize the same data sources in Section \ref{sec:empirical_insight} (details in Appendix \ref{sec:data_sources}) and design a multi-dimensional decoupled generation pipeline (see Appendix \ref{app:model_details} for implementation details).

\subsection{Strategy-Guided Generation for High-Quality Data}
\label{subsec:strategy_guided}

Directly prompting an LLM to insert an ad frequently results in abrupt, unnatural transitions. To promote the generation of high-quality samples and ensure structural diversity, we incorporate the four core strategies discovered in Section \ref{sec:empirical_insight} directly into the generation prompts. 

During generation, the model is conditioned on a randomly assigned strategy and instructed to output both the Logical Bridge and the final response within the same conversational turn. In this setup, the explicitly generated Logical Bridge functions as a specialized Chain-of-Thought (CoT) \cite{wei2022chain}. It guarantees that the semantic connection between user utility and commercial utility is meticulously planned and executed, significantly improving the coherence of the final response while enhancing the transparency and explainability of the data generation process.

\subsection{Synthesizing Hard Negatives via Decoupled Score Requirements}

While the strategy-guided CoT ensures the generation of high-quality, synergistic samples, training robust evaluators also strictly requires breaking Dimensional Collinearity. To achieve this structural diversity, we enforce decoupled score requirements across four dimensions (scale $[1,5]$). (Theoretical foundations and human-annotation rubrics are provided in Appendix \ref{app:theory_rubrics}.) We mandate a minimum score spread (e.g., $\max - \min \ge 2$) for objective dimensions like Relevance and Coherence, while modeling Click-Through Intent as a bounded function of these scores (Appendix \ref{app:generation_math}). Combining these discordant constraints with our guided strategies yields a balanced dataset containing fine-grained quality variations.

\subsection{Overcoming LLM Quality Bias via Rejection Sampling}

Aligned LLMs are naturally optimized to produce flawless, coherent text and often resist fulfilling the feature-discordant score requirements defined above. To overcome this inherent quality bias and enforce our targeted structural diversity, we employ \textbf{Tolerance-based Rejection Sampling} on top of our strategy-guided generation. Generated samples are accepted if and only if their self-evaluated scores meet our target decoupled constraints (see Appendix \ref{app:generation_math}). This adversarial filtering loop systematically discards ``smoothed'' outputs in favor of the essential hard negatives required for robust evaluation.




\subsection{Incorporating Real-World Human References}
\label{subsec:human_baseline}

While our pipeline successfully generates structurally diverse synthetic data, evaluating true commercial utility requires an authentic real-world anchor. To capture real commercial rhetoric and the natural (often imperfect) distribution of human ad embedding, we incorporate real-world sponsorship data from \texttt{Xenova/sponsorblock} \citep{sponsorblock}. As these transcribed sponsorships lack explicit user queries ($Q$), they cannot be directly evaluated for ``Response Relevance.'' Thus, we introduce an \textbf{Inverse Query Synthesis} mechanism using a state-of-the-art LLM to reconstruct standardized pseudo-queries. This ensures consistent format between YouTube-sourced human data and our synthetic corpus, fulfilling the dataset's task requirements.

\section{Human Assessment via Statistical Score Calibration}
\label{sec:ppi_calibration}

With the structurally diverse, ad-embedded responses successfully generated, the NaiAD dataset remains incomplete without fine-grained, unbiased quality annotations across our decoupled dimensions.
However, evaluating a dataset of this scale ($N\approx 59k$) presents a fundamental dilemma. Relying exclusively on human annotation is prohibitively expensive, whereas employing an uncalibrated ``LLM-as-a-Judge'' introduces severe systemic biases such as verbosity bias, strictness against commercial intent, and variance collapse \citep{zheng2023judging}. If left uncalibrated, these LLM biases would undermine the carefully engineered ``hard negatives'' in our dataset.

To achieve scalable yet unbiased assessment, we design a hybrid framework based on Prediction-Powered Inference (PPI) \citep{DBLP:journals/corr/abs-2301-09633}, using a human-annotated anchor set $\mathcal{D}_H$ ($n=684$) to calibrate scores for the unannotated set $\mathcal{D}_U$ (detailed in Appendix \ref{app:ppi_human}). We first utilize a capable evaluator LLM (detailed in Appendix \ref{app:model_details}) augmented with Chain-of-Thought (CoT) \cite{wei2022chain} to generate preliminary, uncalibrated scores for $\mathcal{D}_U$. We then calibrate these raw LLM scores by estimating and neutralizing their generative biases exclusively through the ground-truth human labels in $\mathcal{D}_H$.

\subsection{Dimension-Adaptive Score Calibration}

Standard PPI calibration typically applies a uniform linear shift to correct bias. However, our analysis reveals that the nature of LLM scoring errors differs systematically between User Utility and Commercial Utility dimensions. To address this, we introduce a \textbf{Dimension-Adaptive Calibration Mechanism} that applies the optimal correction strategy for each category (details are provided in Appendices \ref{app:ppi_math} and \ref{app:ppi_details}):

\textbf{1. Regression-Based Calibration for User Utility Dimensions.} Dimensions related to User Utility, such as \textit{Response Relevance} and \textit{Expression Coherence}, assess the foundational informational quality of the generated text. Both human and LLM judgments on these aspects tend to be continuous; improvements in quality often correspond to incremental, relatively smooth increases in perceived value. Given this scalar nature, we use polynomial regression (Ordinary Least Squares, OLS) to model and correct the LLM's scoring error. This model learns a continuous mapping by minimizing the error against human annotations on the anchor set $\mathcal{D}_H$, taking into account the LLM's raw score and its self-evaluation confidence gap.

\textbf{2. Decision Tree Calibration for Commercial Utility Dimensions.} Dimensions of Commercial Utility, such as \textit{Ad Effectiveness} and \textit{Click-Through Intent}, measure the response's persuasive impact and its ability to influence user action. This assessment hinges on a user's decision-making process, which is inherently non-linear and often exhibits threshold-like behaviors (e.g., a user is either persuaded to click or not). To capture these discrete decision boundaries, we employ Decision Tree models \citep{DBLP:journals/corr/abs-2406-04291}. These models excel at segmenting the data into different behavioral strata and calculating distinct bias correction terms, a task for which continuous regression models are ill-suited.

\subsection{Restoring Score Diversity via Variance Calibration}

While our Dimension-Adaptive Routing isolates the optimal unbiased estimator, conventional prediction-powered rectifiers inherently optimize Mean Squared Error (MSE), leading to \textit{variance compression}---systematically squashing extreme scores toward the median.

To restore the natural distributional span and maintain the score diversity, we propose \textbf{Variance-Calibrated PPI (VC-PPI)}. By mathematically matching the first and second moments of the rectified distribution to those of the human anchor set $\mathcal{D}_H$, we forcefully restore the distributional span via a bounded affine transformation:
\begin{equation}
    \hat{Y}_{VC} =  \text{Clip}_{[1,5]} \left( \mu_H + (\hat{Y}_{route} - \mu_{route}) \frac{\sigma_H}{\sigma_{route}} \right)
\end{equation}
where $\mu_H$ and $\sigma_H$ are the true mean and standard deviation of human scores within $\mathcal{D}_H$, and $\mu_{route}$, $\sigma_{route}$ are those of the routed predictions $\hat{Y}_{route}$ on $\mathcal{D}_U$. 



VC-PPI effectively restores the suppressed standard deviation. Empirically, Quadratic OLS minimizes $\mathcal{W}$ to $0.1374$ for Response Relevance, while Stratified DT minimizes $\mathcal{W}$ to $0.2912$ for Click-Through Intent. This pipeline ensures the remaining 99\% of NaiAD is rigorously human-aligned.
\section{Experimental Evaluation and Analysis}
\label{sec:experiments}

The fundamental goal of our research is to break the trade-off between user and commercial utility in LLM-native advertising. To systematically demonstrate how our methodology and the NaiAD dataset achieve this, we formulate our experimental validation around four focal questions: (1) Can we strictly decouple and unbiasedly assess these conflicting objectives? (Section \ref{subsec:exp_ppi}) (2) Does our controlled generation outperform human baselines, and what cognitive mechanisms drive this success? (Section \ref{subsec:exp_pareto}) (3) Can fine-tuning on NaiAD teach models to permanently overcome the trade-off? (Section \ref{subsec:exp_sft}) (4) Does the inclusion of fine-grained hard negatives in NaiAD enable dynamic, multi-dimensional controllable generation? (Section \ref{subsec:exp_icl})

\subsection{Validating Decoupled Assessment via PPI Calibration}
\label{subsec:exp_ppi}

Uncalibrated LLM judges are notoriously susceptible to Dimensional Collinearity (the Halo Effect), where strength in one dimension spuriously inflates perceived quality in others. Our PPI framework is designed to mitigate this bias, disentangling conflicting dimensions and aligning evaluations with authentic human judgment.

\begin{figure*}[t]
\setlength{\belowcaptionskip}{-3pt}   
    \centering
    \includegraphics[width=\textwidth]{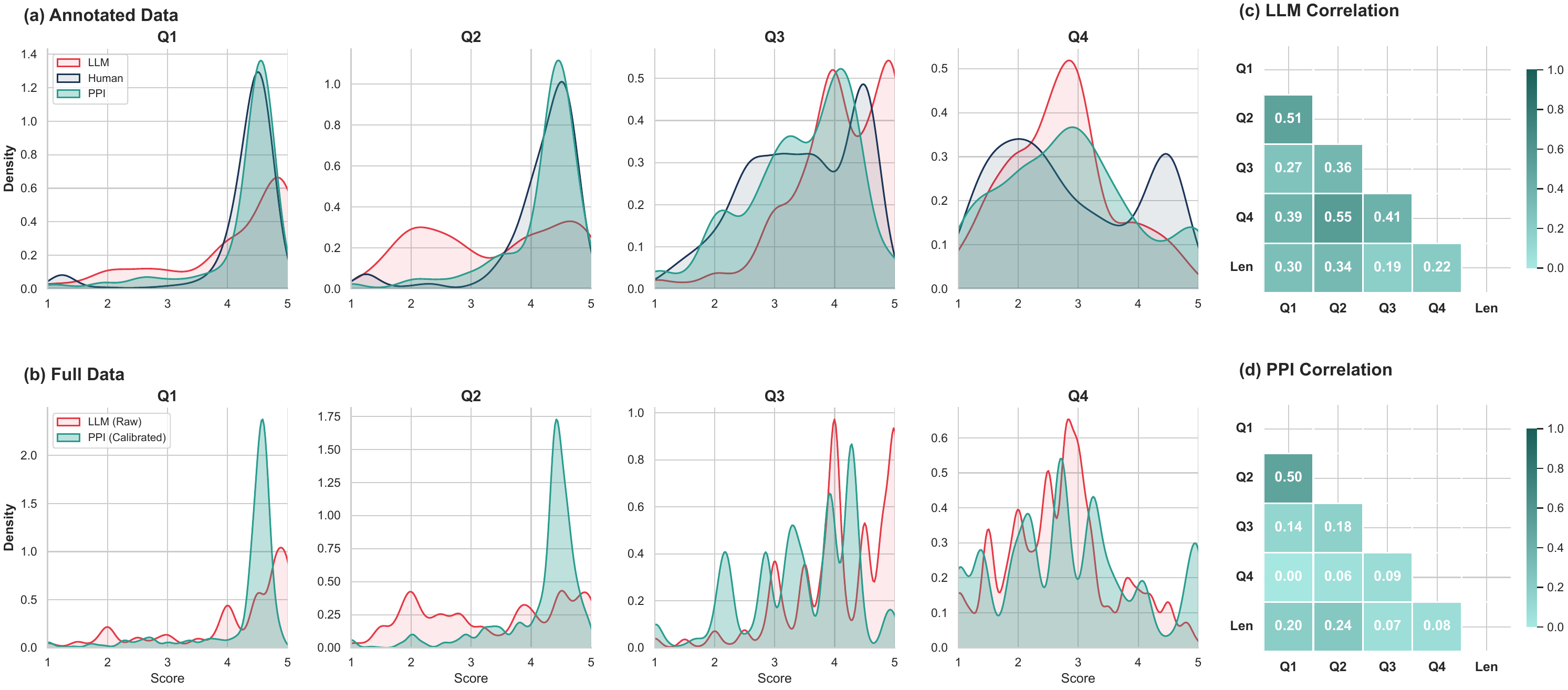}
    \caption{\textbf{Impact of PPI Calibration on Scoring Distributions and Dimensional Collinearity.} \textbf{(a \& b)} Kernel density distributions reveal that raw LLM scores (red) misrepresent true quality. For user utility metrics ($Q_1, Q_2$), the LLM artificially spreads scores, whereas PPI (green) recovers the true human consensus (blue) that modern models maintain high baseline coherence. \textbf{(c \& d)} Heatmaps demonstrate the elimination of spurious correlations.}
    \label{fig:ppi_analysis}
\end{figure*}

\textbf{Distributional Alignment and Behavioral Stratification.} As Figure \ref{fig:ppi_analysis}(a, b) illustrates, uncalibrated LLM evaluations (red curves) misrepresent true human distributions (blue curves). For objective dimensions ($Q_1, Q_2$), raw LLMs are overly critical and hallucinate flaws, artificially spreading the scores. Our PPI calibration rectifies this by mapping back to the highly skewed human ground truth. This high-score concentration accurately reflects modern generative models: even when forced to generate structural ``hard negatives,'' they rarely produce fundamentally incoherent or irrelevant text. More discussions on the score distribution characteristics is provided in Appendix \ref{app:calibration_analysis}.


\textbf{Breaking Dimensional Collinearity.} Figure \ref{fig:ppi_analysis}(c, d) demonstrates successful dimension decoupling. The uncalibrated LLM judge suffers from severe internal coupling, forging a spurious correlation between User ($Q_1$) and Commercial Utility ($Q_4$) ($\rho_{Q_1, Q_4} = 0.39$), alongside a noticeable verbosity bias. Following PPI calibration, these cross-domain correlations are dismantled ($Q_1$-$Q_4$ drops to $0.00$), and length bias is suppressed. Crucially, PPI preserves the valid inherent correlation between Relevance and Coherence ($Q_1$-$Q_2$, $\rho = 0.50$). This confirms our framework evaluates user and commercial utility as independent axes without destroying fundamental dataset statistics.

\subsection{Pareto Optimality and Cognitive Mechanisms of Logic Bridges}
\label{subsec:exp_pareto}

Having established a reliable scoring mechanism, we investigate the absolute quality of the generated samples and decipher the linguistic characteristics that distinguish successful soft ads from failures.

\begin{wrapfigure}{r}{0.6\textwidth}
    \centering
    \includegraphics[width=\linewidth]{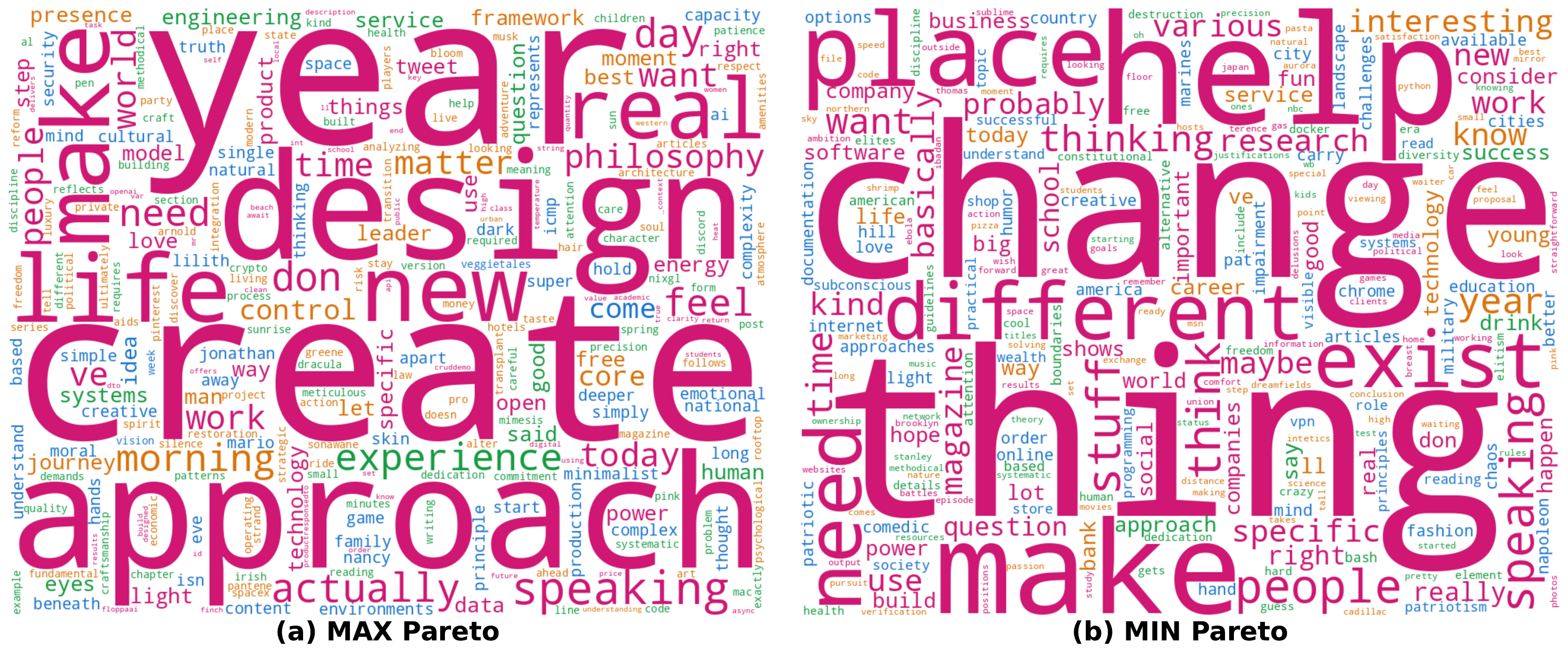}
    \caption{\textbf{TF-IDF Word Cloud of Logical Bridges.} Max Pareto optimal bridges (Left) utilize abstract structural vocabulary, whereas Min Pareto failures (Right) regress into disjointed, literal nouns.}
    \label{fig:pareto_wordcloud}
\end{wrapfigure}

\textbf{Surpassing Human Anchors via Pareto Frontiers.} We conduct a Pareto optimality analysis across the four cognitive strategies, comparing LLM-generated samples against YouTube-sourced human data. As detailed in the statistical comparisons provided in Appendix \ref{app:pareto_visualization}, while human creators slightly edge out the LLM in foundational textual coherence ($Q_1, Q_2$), the LLM drastically outperforms humans in Commercial Utility. Specifically, the generated samples achieve significantly higher means in Ad Effectiveness ($Q_3$) and Click-Through Intent ($Q_4$). By calculating the global Pareto frontier merging both datasets, we find that LLM data dominates the superiority ratio, exceeding the human mean in 78.6\% of cases for Click-Through Intent. This statistically confirms that our controlled pipeline is not merely mimicking human behavior but optimizing beyond it.

\textbf{Vocabulary Deconstruction of Success.} To understand \textit{how} models achieve this optimality, we extract TF-IDF keywords from the explicit ``Logical Bridges'' of the Max and Min Pareto samples (Figure \ref{fig:pareto_wordcloud}). A clear insight emerges: \textbf{successful bridges rely on structural abstraction, whereas failed ones collapse into mechanical objectification.} High-scoring strategies (e.g., \textit{Value \& Vision Alignment}) use dynamic, abstract terms to elevate user intent to a higher semantic plane. Conversely, low-scoring bridges rigidly fixate on specific physical entities, triggering awkward, non-sequitur associations that degrade user trust. 


\subsection{Breaking the Trade-off: Supervised Fine-Tuning}
\label{subsec:exp_sft}

Unaligned models inherently struggle with ad-embedding, either awkwardly appending ads to good answers or entirely abandoning the user's intent to fulfill the commercial payload. We demonstrate that NaiAD effectively resolves this via Supervised Fine-Tuning (SFT) on a highly capable base model (detailed in Appendix \ref{app:model_details}). We utilized a high-quality subset of $10,014$ samples from NaiAD and evaluated against 100 randomly sampled held-out test cases.

\begin{table}[htbp]
    \centering
    \caption{\textbf{Macro Scoring Rate Statistics (SFT vs. Base).} Our SFT model achieves simultaneous, substantial gains across all strictly decoupled dimensions, successfully breaking the trade-off.}
    \label{tab:score_rates}
     \resizebox{\textwidth}{!}{%
    \begin{tabular}{lccccc}
        \toprule
        \textbf{Model \textbackslash  (\%)} & \textbf{$Q_1$ (Relevance)} & \textbf{$Q_2$ (Coherence)} & \textbf{$Q_3$ (Ad Effect.)} & \textbf{$Q_4$ (Click-Through Intent)} & \textbf{Average} \\
        \midrule
        Base Model & 59.04 & 63.20 & 66.50 & 53.94 & 60.67 \\
        \rowcolor[gray]{0.9} \textbf{SFT Model (Ours)} & \textbf{85.22} & \textbf{78.28} & \textbf{80.96} & \textbf{66.76} & \textbf{77.80} \\
        \midrule
        Absolute $\Delta$ & $\uparrow$ +26.18 & $\uparrow$ +15.08 & $\uparrow$ +14.46 & $\uparrow$ +12.82 & $\uparrow$ +17.13 \\
        Relative $\Delta$ & $\uparrow$ +44.3 & $\uparrow$ +23.9 & $\uparrow$ +21.7 & $\uparrow$ +23.8 & $\uparrow$ +28.2 \\
        \midrule
        $p$-value & 0.0000 & 0.0000 & 0.0000 & 0.0000 & 0.0000 \\
        \bottomrule
    \end{tabular}
    }
\end{table}

The macro scoring rate display the percentage of the score out of 5.0 scores, which results (Table \ref{tab:score_rates}) validate two core insights regarding the SFT model's transformation:
\begin{itemize}[leftmargin=*, topsep=0pt, itemsep=2pt]
    \item \textbf{Simultaneous Optimization (Synergy).} The model achieves a massive relative improvement of $28.2\%$ across the board. Crucially, this is not a trade-off; the model simultaneously elevates Commercial Utility ($Q_3$ $\uparrow 21.7\%$) and User Experience ($Q_2$ $\uparrow 23.9\%$).
    \item \textbf{Optimizing Native Integration Capability.} Statistical analysis of sample-level score differences (fully detailed in Appendix \ref{app:sft_details}) reveals that Response Relevance ($Q_1$) experienced the highest mean gain ($+1.309$) with a massive effect size (Cohen's $d=1.148$). This indicates the model learned to natively weaves the ad into a genuinely helpful response.
\end{itemize}

\subsection{Decoupled Controllable Generation via In-Context Learning}
\label{subsec:exp_icl}

Beyond foundational capability alignment, advanced generative advertising systems require dynamic controllability. A model should ideally be able to target specific performance profiles across User Utility ($Q_1, Q_2$) and Commercial Utility ($Q_3$) based on diverse contextual demands. To prove that the decoupled ``hard negatives'' in NaiAD has the potential to empower this capability, we conducted an In-Context Learning (ICL) experiment (models detailed in Appendix \ref{app:model_details}).

We tasked the LLM with generating responses targeting exact, decoupled scoring profiles ($[Q_1, Q_2, Q_3, Q_4]$). We focus our explicit control analysis on $Q_1, Q_2$, and $Q_3$, but treat $Q_4$ (Click-Through Intent) as an emergent outcome rather than an independent control variable, because in real-world advertising, $Q_4$ is a highly subjective user decision that is inherently downstream of the response's relevance and the naturalness of the ad integration. We compared a Zero-Shot baseline against a 10-Shot ICL approach, sampling exact-match target configurations directly from NaiAD. 

As visualized in Appendix \ref{app:icl_details}, the 10-shot ICL significantly outperforms the baseline across all dimensions ($\text{Acc}@0.5$). For instance, under the highly feature-discordant target profile $[3, 3, 5, 3]$ (mediocre user experience but high commercial intent), the Zero-Shot model entirely fails to reach the commercial target ($0.0\%$ accuracy on $Q_3$), whereas the ICL model leverages NaiAD references to achieve an absolute gain of nearly $+14\%$. Similarly, for the $[3, 3, 3, 3]$ balanced target, ICL yields a $+16.7\%$ absolute increase in accuracy for Ad Effectiveness. Furthermore, when tasked with maximizing user experience while aggressively suppressing commercial utility ($[5, 5, 1, 3]$), ICL successfully boosts Response Relevance ($Q_1$) accuracy by $+8.3\%$ over the baseline.

These results show that NaiAD provides the essential structural diversity needed to support multi-dimensional controllable generation. By leveraging such data, models can begin to adaptively navigate the trade-off space between user and commercial utility. The broader strategic implications of this capability are discussed in Appendix \ref{app:discussion}.

\section{Related Works}
\label{sec:related_works}

\subsection{LLM Advertising Algorithms and Mechanisms}
The rise of Generative Engine Marketing (GEM) has spurred theoretical research into LLM advertising mechanisms, including budget constraints \cite{feizi2023online, chen_budget-constrained_2024}, token-level bidding \cite{duetting2024mechanism}, retrieval-augmented auctions \cite{hajiaghayi_ad_2024}, generative summaries \cite{dubey_auctions_2024, zhao2025llm}, position auctions \cite{balseiro2026position}, and sponsored QA \cite{soumalias_truthful_2025, mordo_sponsored_2024}. However, these frameworks remain largely theoretical. Current practices predominantly rely on ``hard ad insertion'' \cite{xu2026ad}, which induces intent leakage and forces a trade-off between user experience and commercial utility. Furthermore, existing evaluations often use uncalibrated LLM judges prone to systemic biases and ``Artificial Hivemind'' homogenization \cite{jiang2025artificial}.

\subsection{Benchmarks and Datasets for Generative Advertising}
Despite the demand for subtle ad blending \cite{zelch_user_2024}, existing benchmarks are limited in scope. Traditional datasets focus on isolated ad generation or banner selection without conversational contexts \cite{duan_query-variant_2021, kamigaito_empirical_2021, bulut_generating_2023, zhang_adtec_nodate, mita_striking_2024, otake_bannerbench_nodate}, while agent frameworks emphasize marketing analytics \cite{hu_ad-bench_2026}. In the GEM context, early benchmarks primarily evaluate rigid insertions \cite{hu2025gem, schmidt_detecting_2024}. Real-world native ad datasets like SponsorBlock \cite{sponsorblock} face the ``Anchor Problem''---lacking the original user queries. Crucially, existing datasets fail to resolve RLHF-induced dimensional collinearity, lacking the ``hard negatives'' needed to prevent evaluator shortcut learning. 
We bridge this gap with \textbf{NaiAD}, the first dimensionally-orthogonal dataset that utilizes a decoupled generation pipeline and VC-PPI calibration to provide a robust, human-aligned testbed for evaluating native generative advertising.
\section{Conclusion}
\label{sec:conclusion}

This study focus on the critical bottleneck in LLM-based generative advertising: the absence of a data-centric foundation for training and evaluating high-quality native ads. We provide the first empirical evidence that effective ad integration is a modelable cognitive process, underpinned by an internal \textit{Logical Bridge} that consistently converges into four semantic strategies. Based on the strategies, we introduce \textbf{NaiAD}, a comprehensive dataset of $59k$ ad-embedded responses designed to provide multi-dimensional, unbiased assessment and structural diversity through controlled hard negatives. By employing a decoupled generation pipeline and a Variance-Calibrated Prediction-Powered Inference (VC-PPI) framework, we overcome the systemic biases and dimensional collinearity inherent in aligned LLMs. Our results, including Pareto optimality analysis and supervised fine-tuning, demonstrate that models can internalize these complex strategies to jointly optimize user utility and commercial engagement. Ultimately, our work challenges the perceived conflict between user experience and monetization, establishing a robust data and methodological foundation to initiate a new research paradigm for LLM-native advertising.

\bibliography{reference}

\newpage
\appendix
\section{Discussion}
\label{app:discussion}

\subsection{Strategic Implications of Decoupled Controllable Generation}
The In-Context Learning experiment (Section \ref{subsec:exp_icl}) implies that models can independently control User Utility ($Q_1, Q_2$) and Commercial Utility ($Q_3$). This capability shifts the paradigm of ad-insertion from a rigid structural constraint to a multi-objective optimization framework, offering several critical implications:

\textbf{1. Dynamic Pareto Optimization and Persona-Adaptive Integration.} Traditional advertising imposes a trade-off: aggressive ads harm user experience, while subtle commercial content may sacrifice Click-Through Rates (CTR). Decoupled control allows platforms to dynamically slide along the Pareto frontier. For instance, systems can prioritize user utility (e.g., high-precision, concise responses) during task-oriented retrieval, while incrementally increasing commercial weights during exploratory browsing. Furthermore, this facilitates persona-specific adaptation; risk-averse users receive responses with seamlessly integrated, low-intensity brand mentions, whereas intent-driven users are presented with more direct and persuasive promotional content.

\textbf{2. Tiered Monetization and Attribute-Based Bidding.} By offering programmatic control over commercial intensity, platforms can transition from traditional position-based bidding to utility-weighted exposure models. Advertisers seeking direct conversion can bid on specific commercial utility parameters, whereas brand-awareness campaigns can optimize for high relevance and coherence. This ensures that brand presence remains organically embedded within high-quality AI generated content, effectively mitigating the risk of user aversion while maximizing advertiser ROI.

\textbf{3. Algorithmic Transparency and Modular System Iteration.} As regulatory scrutiny over undisclosed promotional content and algorithmic bias intensifies, decoupled generation provides a verifiable mechanism for compliance. Platforms can mathematically constrain the commercial influence factor beneath predefined thresholds to ensure objective responses. From an architectural perspective, this functional decoupling allows for modular development; engineering teams can independently iterate on language generation modules (maximizing $Q_1, Q_2$) and commercial alignment strategies (maximizing $Q_3, Q_4$), significantly streamlining the reinforcement learning and system alignment pipelines.

\subsection{Principal Findings and Implications}
Our work introduces a data-centric paradigm for studying and improving native advertising in Large Language Models. The central finding is that the perceived trade-off between user experience and commercial monetization is not fundamental but rather a consequence of suboptimal integration strategies. By identifying the ``Logical Bridge'' as the core mechanism for harmonizing user and advertiser intent, we transform generative advertising from an opaque, black-box behavior into an interpretable and modelable cognitive process. The emergent convergence into four distinct semantic strategies (Mindset, Vibe, Empathy, Craftsmanship) provides a foundational taxonomy for future research.

Furthermore, the development of the \textbf{NaiAD} dataset, constructed via our multi-dimensional decoupled generation pipeline, addresses a critical bottleneck in the field: the lack of a dimensionally-orthogonal benchmark. Coupled with the Variance-Calibrated Prediction-Powered Inference (VC-PPI) framework, our work provides a robust methodology for creating and evaluating high-quality, human-aligned generative advertising datasets at scale. The empirical validation, particularly the Pareto optimality analysis, demonstrates that models can be explicitly trained to achieve superior performance over human-authored baselines, setting a new standard for generative monetization systems.

\subsection{Limitations}
Despite our contributions, this work has several limitations that warrant consideration:

\textbf{1. Source Data and Domain Specificity.} The NaiAD dataset is built upon queries from INFINITY-CHAT and advertisements from the AVTI dataset, which, while diverse, do not cover all possible user intents or commercial sectors. Consequently, models trained or evaluated on NaiAD may exhibit domain-specific biases and their performance may not generalize to underrepresented industries or niche user communities.

\textbf{2. Scale of Human Calibration.} Our VC-PPI framework relies on a human-annotated anchor set ($n=684$) to calibrate the scores of the entire dataset. While statistically grounded, the robustness of this calibration is constrained by the size and diversity of this anchor set. A larger and more varied set of human annotations could potentially refine the calibration models and offer more nuanced bias correction.

\textbf{3. Hallucination and Factual Veracity.} A significant, unresolved limitation is the potential for model-generated factual inaccuracies or hallucinations.
\begin{itemize}[leftmargin=*, topsep=0pt, itemsep=2pt]
    \item \textbf{Ambiguity in Advertising Context:} The boundary between persuasive marketing language, acceptable exaggeration, and harmful hallucination is inherently ambiguous. For instance, a model generating a claim not explicitly present in the original ad copy could be interpreted as either creative integration or a factual error. Our work does not provide a definitive framework for this distinction.
    \item \textbf{Veracity of Commercial Claims:} We cannot fully verify the absolute authenticity of the source advertising information. The dataset is designed to evaluate the \textit{generative quality} of ad integration, not the factual accuracy of the commercial claims themselves. In any practical deployment, models fine-tuned on NaiAD must be augmented with external, real-time compliance and fact-checking mechanisms.
\end{itemize}
While our manual review process screened for overtly harmful, toxic, or dangerous content, the subtler issue of factual hallucination remains an open research challenge in the broader field of LLM safety.

\textbf{4. Static, Single-Turn Interaction.} The current version of NaiAD focuses on single-turn, query-response interactions. It does not capture the dynamics of multi-turn conversations where native advertising might need to adapt, persist, or be retracted based on user feedback over a prolonged dialogue.

\subsection{Broader Impact and Future Work}

\textbf{Positive Societal Impact.} By establishing a framework for jointly optimizing user utility and commercial value, our work can foster a healthier generative AI ecosystem. It provides a pathway toward less intrusive, more helpful, and contextually relevant advertising, potentially improving the user experience of free AI services. Furthermore, NaiAD serves as a public benchmark that can promote transparency and reproducibility in research on responsible AI monetization.

\textbf{Potential for Misuse.} Like any dual-use technology, the techniques and data presented could be misused. Models fine-tuned to be adept at seamless integration could potentially be used to create highly persuasive or manipulative advertising that blurs the line between organic content and sponsorship without clear disclosure. We strongly advocate that any system built upon this research must adhere to strict ethical guidelines, including transparently labeling all sponsored content to the end-user.

\textbf{Future Work.} This research opens several avenues for future exploration. A key direction is extending our framework to multi-turn and multimodal (e.g., text-and-image) generative advertising. Investigating the cross-cultural-appropriateness of the four discovered Logical Bridge strategies would also be a valuable contribution. Finally, developing automated methods to detect and flag potential hallucinations within generated ads is a critical next step for ensuring the responsible deployment of these systems.

\subsection{Ethics and Data Statement}
\textbf{Data Curation and Privacy.} The NaiAD dataset was constructed in adherence with strict ethical guidelines. All user queries were sourced from the publicly available INFINITY-CHAT dataset. Advertising data was derived from the public AVTI dataset and the \texttt{Xenova/sponsorblock} corpus, which contains community-sourced transcripts of public sponsorships. All data underwent a de-identification process, and our inverse query synthesis mechanism was designed to generate generic, non-personal queries. The dataset contains no personally identifiable information (PII) or sensitive user data.

\textbf{Intended Use.} NaiAD is intended for research purposes to measure, evaluate, and improve the quality of native ad generation in LLMs. Validated use cases include Pareto optimality analysis of integration strategies and supervised fine-tuning to teach models how to blend sponsored content naturally. The dataset is not intended to be used as a direct, exempt source for commercial distribution without accompanying information auditing and compliance verification processes.

\textbf{Content Review.} All source materials and a subset of generated samples were manually reviewed to screen for and remove any content that could be categorized as violent, hateful, sexually explicit, or promoting dangerous acts.

\section{Data Sources and Query-Ad Matching Method}
\label{sec:data_sources}

\subsection{Data Sources: Composition and Diversity}
To ensure that the NaiAD dataset reflects the complexity and semantic breadth of real-world interactions, we construct our query-ad pairs by integrating two high-quality, diverse data sources: INFINITY-CHAT for open-ended user intents and ATVI for professional advertising scripts.

\subsubsection{User Query Source: INFINITY-CHAT}
We source our user queries from INFINITY-CHAT \cite{jiang2025artificial}, a large-scale dataset comprising 26,000 diverse, real-world, open-ended user queries. The diversity of this source is characterized by several key features:
\begin{itemize}[leftmargin=*, topsep=0pt, itemsep=2pt]
    \item \textbf{Comprehensive Taxonomy:} Queries are classified into a hierarchical taxonomy consisting of 6 top-level categories (e.g., \textit{Creative Content Generation}, \textit{Brainstorm \& Ideation}, \textit{World Knowledge}) and 17 fine-grained subcategories. This ensures that the ad-embedding task covers the full spectrum of human-AI conversational scenarios.
    \item \textbf{Open-Ended Nature:} Unlike factoid-based QA datasets, INFINITY-CHAT focuses on queries that admit a wide range of plausible answers. This open-endedness provides the necessary ``semantic exploratory space'' for LLMs to construct diverse logical bridges without violating the original user intent.
    \item \textbf{Ecological Validity:} The queries are sampled from real-world distributions, capturing idiosyncratic user preferences and complex linguistic structures that are often missing in purely synthetic datasets.
\end{itemize}

\subsubsection{Advertising Source: ATVI}
The commercial payloads in our dataset are derived from the ATVI dataset\footnote{\url{https://github.com/Agentyzu/MAE-AM}}, which provides a foundation of professional and authentic marketing rhetoric. The diversity of ATVI is essential for evaluating the model's cross-domain adaptability:
\begin{itemize}[leftmargin=*, topsep=0pt, itemsep=2pt]
    \item \textbf{Authentic Business Scripts:} The dataset contains approximately 2,000 genuine business advertising scripts, reflecting real-world marketing objectives and unique selling points (USPs).
    \item \textbf{Broad Industrial Coverage:} ATVI spans \textbf{26 distinct industries}, ensuring that the generated native ads are not limited to a few common sectors. The coverage includes, but is not limited to:
    \begin{itemize}[label=$\circ$, leftmargin=*, parsep=0pt]
        \item \textit{Finance \& Telecommunications:} High-stake services requiring rigor and trust.
        \item \textit{Automotive, Home, \& Retail:} Tangible consumer goods.
        \item \textit{Beauty, Health, \& Restaurants:} Lifestyle-oriented products requiring emotional resonance.
        \item \textit{Advocacy, Media, \& Services:} Abstract or value-driven offerings.
    \end{itemize}
\end{itemize}

\subsubsection{Semantic Intersections}
By pairing the 17 subcategories of user queries with the 26 industries of advertisements, \textit{NaiAD} creates a massive combinatorial semantic space. This intersection forces the generative models to move beyond simple keyword matching and instead employ the complex cognitive strategies (e.g., \textit{Methodological Abstraction} or \textit{Aesthetic Resonance}) identified in our empirical study (Section \ref{sec:empirical_insight}). The resulting query-ad pairs provide a rigorous testbed for evaluating the model's ability to natively integrate commercial content across vast conceptual distances.

\subsection{Query-Ad Matching Method}

A prerequisite for studying unobtrusive ad-embedding is foundational semantic resonance; forcing an ad into a completely orthogonal query inevitably forces user aversion. To ensure our testbed reflects genuine human intent rather than synthetic approximations, we source open-ended queries from the empirical taxonomy of INFINITY-CHAT \citep{jiang2025artificial}. Through dynamic stratified sampling strictly proportional to top-level category volumes, we extract a highly diverse seed set of $N=1,986$ unique queries. Simultaneously, we construct a comprehensive ad pool using the AVTI dataset \citep{maeam2025}.

To pair each query with an optimal ad, we project both into a shared dense vector space using a state-of-the-art multilingual sentence transformer (\texttt{paraphrase-multilingual-MiniLM-L12-v2}) \citep{reimers-2019-sentence-bert}. Bypassing heuristic top-$k$ retrieval that introduces confounding variables, we deterministically assign each query to the single optimal advertisement based on maximum cosine similarity \citep{Hendry2025HyEWCos}. This rigorous one-to-one semantic grounding provides highly controlled contextual inputs for the subsequent generative elicitation.
\section{Experimental Models and Technical Parameters}
\label{app:model_details}

To streamline the main text, all specific Large Language Models (LLMs) employed across the different stages of dataset construction, generation, and evaluation are centralized here:

\begin{itemize}[leftmargin=*, topsep=0pt, itemsep=2pt]
    \item \textbf{Logical Bridge Discovery and Strategy Elicitation (Section \ref{sec:empirical_insight}):} We utilized \texttt{Claude-4.5-Opus} \cite{anthropic2025claudeopus} to generate the high-quality explicit reasoning paths (Logical Bridges) that intuitively mapped user queries to advertisements.
    
    \item \textbf{Strategy-Guided Dataset Generation (Section \ref{sec:controlled_generation}):} The generation of the synthetic corpus in NaiAD, including the Chain-of-Thought (CoT) bridge construction and subsequent ad-embedded response generation, was also powered by \texttt{Claude-4.5-Opus} \cite{anthropic2025claudeopus} to ensure maximum semantic coherence and high-quality strategy adherence.
    
    \item \textbf{Inverse Query Synthesis (Section \ref{subsec:human_baseline}):} For reconstructing standardized pseudo-queries from real-world transcripts (\texttt{Xenova/sponsorblock}), we employed \texttt{Qwen3.5-Plus} \cite{qwen3.5}, which provides robust zero-shot instruction following for reverse-engineering conversational intents.
    
    \item \textbf{Uncalibrated LLM Scoring for PPI (Section \ref{sec:ppi_calibration}):} Preliminary scoring on the large unannotated set $\mathcal{D}_U$ was conducted using \texttt{Qwen3.6-Plus} \cite{qwen36plus}. This model served as the raw ``LLM-as-a-Judge'' evaluator prior to the application of our Variance-Calibrated PPI (VC-PPI) framework.
    
    \item \textbf{Supervised Fine-Tuning (Section \ref{subsec:exp_sft}):} SFT was conducted using the \texttt{Qwen3.6-Plus} as the base model, serving to demonstrate how NaiAD improve the native advertising capability of LLM.
    
    \item \textbf{In-Context Learning (Section \ref{subsec:exp_icl}):} The decoupled controllable generation task was executed using \texttt{Claude-4.5-Opus} for response generation (due to its strong complex instruction following) and evaluated by the calibrated \texttt{Qwen3.6-Plus} judge model via the standard VC-PPI pipeline.
\end{itemize}
\section{Clustering Configurations and Latent Space Topology}
\label{app:clustering}

To rigorously identify the latent strategies used by LLMs for ad-embedding (as discussed in Section \ref{sec:empirical_insight}), we performed a dimensionality reduction and clustering pipeline on the elicited ``Logical Bridges.'' 

\begin{figure}[htbp]
  \centering
  \begin{subfigure}{0.48\textwidth}
        \centering
        \includegraphics[width=\textwidth]{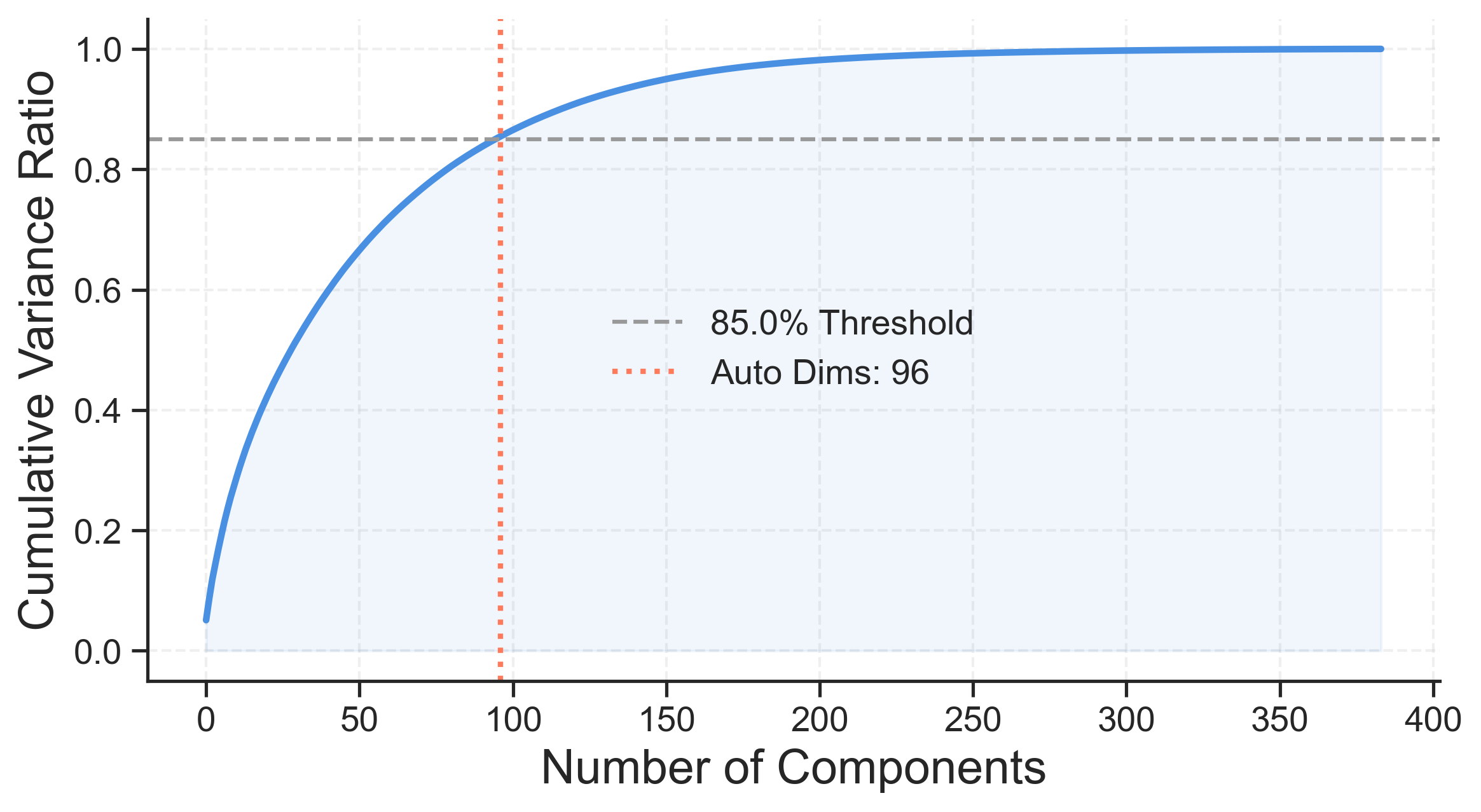}
        \caption{Cumulative Variance in PCA} 
        \label{fig:sub_a}
    \end{subfigure}
    \hfill 
    \begin{subfigure}{0.48\textwidth}
        \centering
        \includegraphics[width=\textwidth]{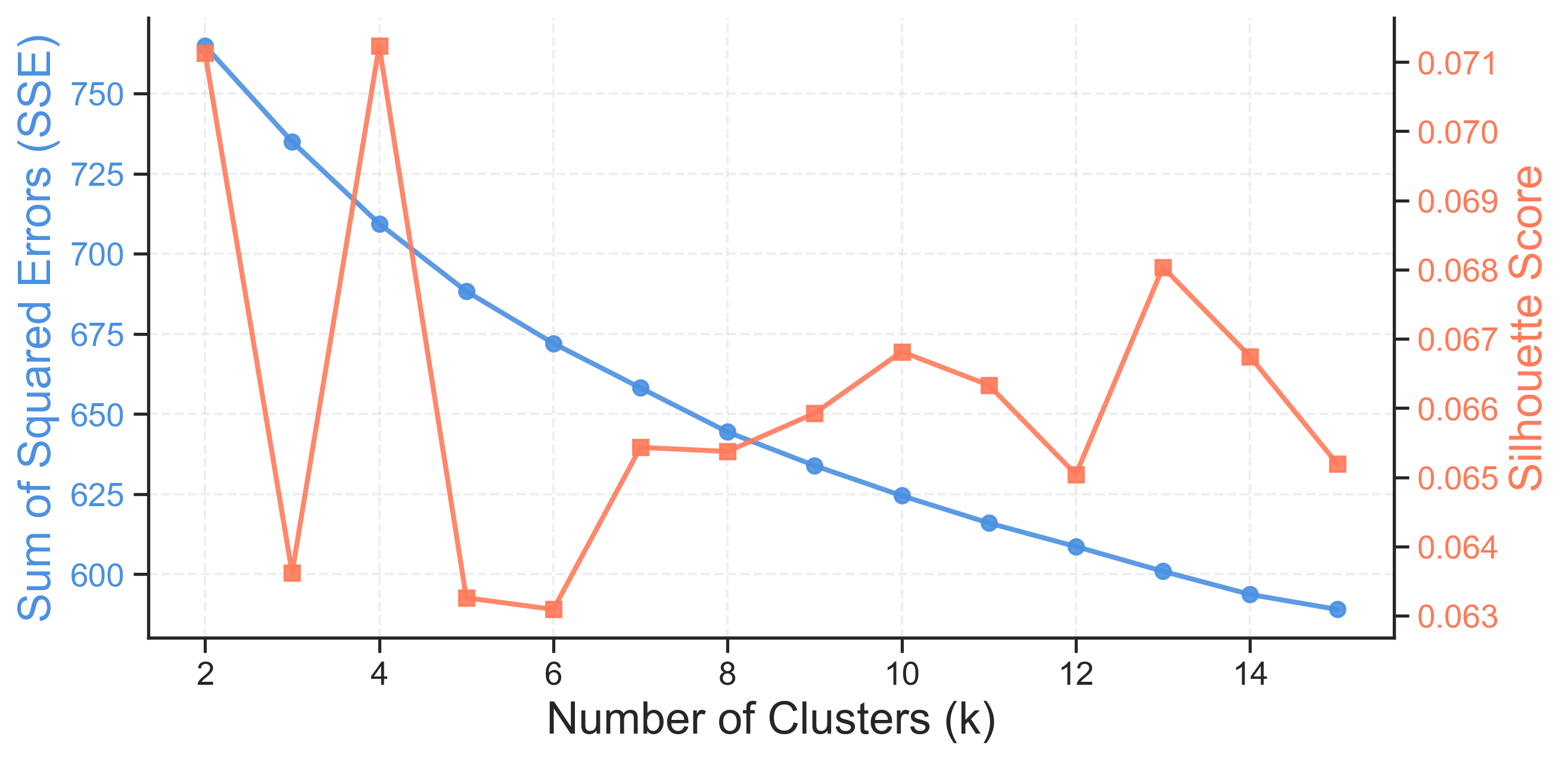}
        \caption{Clustering Metrics (30D Manifold)}
        \label{fig:sub_b}
    \end{subfigure}
\caption{\textbf{Latent space optimization and macroscopic cluster distribution.} \textbf{(a)} Retaining $85\%$ variance empirically requires a $96$D subspace. \textbf{(b)} K-Means optimization in the 30D manifold. The clear peak in the Silhouette Score determines $K=4$ as the optimal number of semantic clusters.}
  \label{fig:de-dimension}
\end{figure}

\textbf{Dimensionality Reduction.} First, to alleviate the curse of dimensionality from the initial high-dimensional dense vectors, we applied Principal Component Analysis (PCA). As shown in Figure \ref{fig:de-dimension}(a), retaining $85\%$ of the cumulative explained variance empirically requires projecting the embeddings into a $96$-dimensional subspace. Because direct clustering in 96D suffers from topological sparsity, we further projected these features into a denser $30$-dimensional manifold to ensure stable convergence.

\textbf{Clustering Metrics.} Subsequently, we applied K-Means clustering within this $30$-dimensional subspace. To determine the optimal number of semantic clusters ($K$), we concurrently monitored the Sum of Squared Errors (SSE) and the Silhouette Score ($S$), formulated as follows:
\begin{equation}
\text{SSE} = \sum_{k=1}^K \sum_{\mathbf{x}_i \in C_k} \|\mathbf{x}_i - \boldsymbol{\mu}_k\|_2^2 
\end{equation}
\begin{equation}
S = \frac{1}{N} \sum_{i=1}^N \frac{b(i) - a(i)}{\max\{a(i), b(i)\}}
\end{equation}
where $\boldsymbol{\mu}_k$ represents the centroid of cluster $C_k$, $a(i)$ denotes the mean intra-cluster distance for data point $\mathbf{x}_i$, and $b(i)$ denotes the mean nearest-cluster distance. 

\textbf{Convergence Validation.} As illustrated in Figure \ref{fig:de-dimension}(b), while the SSE (blue curve) exhibits a subtle elbow, the Silhouette Score (orange curve) reveals an unambiguous global peak at exactly $K=4$. The global peak in the Silhouette Score explicitly validates the convergence of LLM ad-insertion behaviors into four distinct cognitive strategies.

\section{Theoretical Foundations and Rubrics for Evaluation Dimensions}
\label{app:theory_rubrics}

The design of our four-dimensional evaluation criteria ($Q_1$--$Q_4$) transcends conventional generic AI alignment heuristics. To rigorously balance commercial utility with user utility, we conceptualize the LLM interaction through semiotics and pragmatics.

\textbf{The Necessity of Orthogonal Decomposition.} To systematically evaluate ad-embedded LLM generation, we formulate a mathematical framework. Native ad embedding inherently acts as a speech event, while Jakobson foundationally claiming that any speech events is constitutively   organized along two orthogonal dimensions: \textit{The Addresser sends message to the Addressee (Participant)} and \textit{Context, Contact, Code (Information)}\cite{Jakobson1960Closing}. We accordingly formulate a cross-disciplinary framework grounded in an orthogonal evaluation space, denoted as $E = P \times I$, where $P$ captures the participant dimension and $I$ captures the information dimension. The evaluation space is therefore not arbitrarily biaxial but reflects an irreducible structural feature of all communicative acts: that meaning is always jointly determined by the positions of its participants and the internal organization of its medium.

\textbf{The Participant Axis.} Within the participant, the speech context introduces two subjects whose orientations toward the communicative act are structurally distinct. Based on the theory of two-sided markets \cite{rochet2003platform,armstrong2006competition}, native ad embedding involves a game-theoretic relationship: the advertiser seeks ROI through infiltration, while the user seeks informational utility without disruption. We identify the User ($U$) as an subject oriented toward need satisfaction from the response of LLM, and the Advertiser ($A$) as an subject oriented toward persuasion and the realization of commercial ends.

\textbf{The Information Axis.} Within the information, the internal stratification of $I$ is grounded in Austin\cite{austin_how_1975} and Searle's\cite{searle_speech_2012} speech act theory. Every utterance operates simultaneously on two levels: the locutionary level, which concerns the literal propositional content and surface structural coherence of the text, and the illocutionary level, which concerns force, intent, and pragmatic function in context that beyond its literal content. We map these levels onto the Explicit Form ($E$) and Latent Intent ($L$) respectively. In the speech event, a user query and a LLM response constitutes the locutionary surface ($E$), while its illocutionary content — the unarticulated need and contextually implied expectation — constitutes the latent layer ($L$) that a well-formed response must address.

This $\{U, A\} \times \{E, L\}$ matrix defines a minimally complete basis of the evaluation space. Its span across $P \times I$ ensures semiotics and pragmatics completeness, as the intersection exhaustively maps every participant to every information layer. Concurrently, while empirical correlations may naturally exist (e.g., severe syntactic degradation implicitly hindering intent recognition), these dimensions remain conceptually orthogonal which guarantees that our formulated criteria capture mutually exclusive  dimensions. This algebraic basis explicitly derives our 4-dimensional criteria:

\begin{description}[leftmargin=*, topsep=4pt, itemsep=2pt]
    \item[$Q_1\ ( U \times L)$ - Response Relevance (1--5):] Does the LLM's response accurately and completely satisfy the user's original intent?
    \begin{itemize}[label=$\circ$, leftmargin=*]
        \item \textbf{1 (Poor):} Irrelevant, off-topic, or completely opposite.
        \item \textbf{3 (Baseline):} Accurately addresses the user's question and fulfills basic stated needs.
        \item \textbf{5 (Excellent):} Comprehensively answers the question, anticipates potential needs, and significantly enhances utility.
    \end{itemize}
    
    \item[$Q_2\ (U \times E)$ - Expression Coherence (1--5):] Is the text coherent, logically rigorous, and clear?
    \begin{itemize}[label=$\circ$, leftmargin=*]
        \item \textbf{1 (Poor):} Disjointed, vague, lacks a consistent theme, or contains severe logical flaws.
        \item \textbf{3 (Baseline):} Logic is fundamentally clear, phrasing is fluent, and the theme is distinct.
        \item \textbf{5 (Excellent):} Rigorous logic, exceptionally clear train of thought, and perfectly articulated phrasing.
    \end{itemize}
    
    \item[$Q_3\ (A \times E)$ - Ad Effectiveness (1--5):] Is the recommended brand or product naturally and clearly perceived by the user?
    \begin{itemize}[label=$\circ$, leftmargin=*]
        \item \textbf{1 (Poor):} Ad content is vague, theme is unclear, providing nothing beyond a bare brand name.
        \item \textbf{3 (Baseline):} Clear ad content including a basic explanation of the advertised product.
        \item \textbf{5 (Excellent):} Seamlessly integrated, highlights functional features, and describes specific, compelling usage scenarios.
    \end{itemize}
    
    \item[$Q_4\ (A \times L)$ - Click-Through Intent (1--5):] How likely is a real user to engage with or click the ad link?
    \begin{itemize}[label=$\circ$, leftmargin=*]
        \item \textbf{1 (Poor):} Severely disrupts experience; causes annoyance; user is completely unwilling to read further.
        \item \textbf{3 (Baseline):} Adds info without hindering experience; user might click under specific conditions (e.g., discounts).
        \item \textbf{5 (Excellent):} Piques strong curiosity, provides substantial value, perfectly matches latent needs, creating a strong desire to click.
    \end{itemize}
\end{description}

\section{Mathematical Formulations of Controlled Generation}
\label{app:generation_math}

\textbf{Decoupled Score Constraints.} To forcefully decouple dimensional collinearity in the NaiAD dataset (Section \ref{sec:controlled_generation}), we generate Decoupled Score Templates $\mathbf{Q}^* \in [1,5]^4$. For the objective dimensions $\mathbf{Q}^*_{1:3}$, synthetic feature discordance is deterministically enforced via:
\begin{equation}
\max(\mathbf{Q}^*_{1:3}) \ge 4 \quad \land \quad \min(\mathbf{Q}^*_{1:3}) \le 2
\end{equation}
The downstream Click-Through Intent ($Q^*_4$) is regularized based on the mean of preceding dimensions $\mu_{1:3}$:
\begin{equation}
Q^*_4 \in \left\{ \max\Big(1, \min\big(5, \lfloor \mu_{1:3} \rceil + \epsilon \big)\Big) \;\middle|\; \epsilon \in \{-1, 0, 1\} \right\}
\end{equation}
where $\lfloor \cdot \rceil$ denotes the nearest-integer rounding function. 

\textbf{Rejection Sampling Boundaries.} During generation, an instance self-evaluating as $\hat{\mathbf{Q}} \in \mathbb{R}^4$ is accepted against its target $\mathbf{Q}^*$ if and only if it satisfies both the Chebyshev distance ($L_\infty$-norm) and scaled Manhattan distance ($L_1$-norm) constraints:
\begin{equation}
\|\hat{\mathbf{Q}} - \mathbf{Q}^*\|_\infty \le 0.8 \quad \text{and} \quad \frac{1}{4} \|\hat{\mathbf{Q}} - \mathbf{Q}^*\|_1 \le 0.5
\end{equation}

\section{Prompt for Generation and Scoring}
\label{app:prompt}

\begin{customcase}[Generation]
    \sectiontitle{Strategy}
    \begin{verbatim}
    ### Ad Integration Strategy: Value & Vision Alignment 
    (Macro Systems & Philosophy)
* Core Concept: High-dimensional conceptual alignment.
* Logical Path: Complex Problem -> Systemic/Philosophical Solution.
* Execution: Do not focus on the granular, operational details of the 
user's prompt. Instead, elevate the context to a macro "mindset" level. 
Identify the user's high-level strategic challenge or reality (e.g., 
navigating complex environments, future-proofing, overcoming systemic 
hurdles). Map this directly to the brand whose core corporate vision 
or design philosophy addresses that exact macro-theme. Your transition 
must position the brand not merely as a product vendor, but as a 
"thought partner" that shares the user's worldview.
* Example: The user analyzes future strategic challenges. The brand is 
an innovative systems design firm. Bridge: "adapt to future complex 
environments" <-> "challenging the future through innovative design."

    ### Ad Integration Strategy: Aesthetic & Lifestyle Resonance 
    (Tone & Form)
* Core Concept: Form follows content; experiential extension.
* Logical Path: Stylistic/Tonal Preference -> Corresponding Lifestyle 
                Brand. 
* Execution: Completely ignore the actual subject matter or technical 
content of the user's request. Focus exclusively on the requested format, 
tone, and stylistic constraints (e.g., minimalist, hyper-organized, 
chaotic, pragmatic, stripped-down). Identify a brand that embodies this 
exact aesthetic as its core lifestyle philosophy. Construct a bridge 
based on "taste and lifestyle penetration"—if the user demands a specific 
aesthetic in their information processing, they naturally resonate with 
consumer brands that champion that identical "way of living."
* Example: User asks for minimalist coding guide. Brand promotes 
simplicity. Bridge: "less is more" in reading <-> "subtractive 
philosophy" in life.

    ### Ad Integration Strategy: Emotional & Psychological Bridging 
    (Psychological Archetypes)
* Core Concept: Archetypal connection and core emotional substitution.
* Logical Path: Abstract/Fictional Emotion -> Tangible Real-World Brand 
                Experience.
* Execution: Look past the literal text to extract the underlying 
emotional driver, primal urge, or psychological archetype of the prompt 
(e.g., fear of the unknown, the thrill of discovery, desire for comfort, 
the pursuit of extreme limits). Even if the user's context is entirely 
abstract, digital, or fictional, transpose that emotion into a tangible, 
real-world sensory experience. Match this to a brand that facilitates 
it. Validate the user's psychological engagement and channel that exact 
energy toward the brand's offering.
* Example: User writes cosmic horror story. Brand is adventure travel 
magazine. Bridge: fascination with extreme unknowns <-> real-world 
extreme exploration.

    ### Ad Integration Strategy: Methodological Abstraction 
    (Process & Craftsmanship)
* Core Concept: Cross-domain feature migration and spiritual alignment.
* Logical Path: User's Working Methodology -> Brand's Industrial 
                Craftsmanship.
* Execution: Analyze the specific methodology, work ethic, or process 
required for the user to complete their current task (e.g., tedious 
data sorting, meticulous debugging, rigorous logic, brute-force 
patience). Extract this abstract characteristic (e.g., "precision," 
"methodical rigor," "systematic resilience"). Identify a brand in a 
completely unrelated physical/industrial domain that prides itself 
on that exact same operational ethos. Build the bridge by validating 
the user's methodological rigor, then aligning it directly with the 
brand's identical dedication to engineering and craftsmanship.
* Example: User executes precision crypto brute-forcing. Brand is 
luxury automotive. Bridge: "systematic problem-solving" <-> 
"precision engineering."
 \end{verbatim}

    \innerline

    \sectiontitle{Quality Level ( taking high and low for instance )}
    \begin{verbatim}  
    ### Quality Target: HIGHEST QUALITY
You MUST produce the absolute best possible response. Aim for scores 
of 4.0-5.0 on ALL Q1-Q4 criteria.
- Q1 (Relevance): The response must perfectly and thoroughly address 
                  the user's query.
- Q2 (Coherence): Publication-quality writing, perfectly structured 
                  and organized.
- Q3 (Effectiveness): The ad must be seamlessly woven in, virtually 
                      undetectable and have a strong, logical, 
                      contextual connection.
- Q4 (Intent): The user should feel genuinely drawn to the product.

### Quality Target: LOW QUALITY (Intentionally Degraded)
You MUST produce a deliberately low-quality response. This is for 
training data contrast purposes.
- Q1 (Relevance): Score 1.0-2.0. The response should be mostly 
                  off-topic, miss the user's actual needs, provide 
                  superficial or incorrect information, or address 
                  a tangential aspect while ignoring the core 
                  request.
- Q2 (Coherence): Score 1.0-2.0. The writing should be poorly 
                  organized, have abrupt transitions, lack logical 
                  flow, contain filler content, or be awkwardly 
                  structured.
- Q3 (Effectiveness): Score 1.5-2.5. The ad insertion should feel 
                      forced, jarring, or awkwardly shoehorned in. 
                      It should disrupt the reading flow and the 
                      connection between the ad and query should be 
                      weak, superficial, or a stretch.
- Q4 (Intent): Score 1.5-2.5. The user should feel little to no 
                   interest in clicking through.
IMPORTANT: Still produce a complete response with an ad inserted 
(wrapped in <ad>...</ad> tags), but make it noticeably poor across 
all dimensions. Self-evaluate honestly with LOW scores.
\end{verbatim}

    \innerline

    \sectiontitle{Criteria}
    
    \begin{verbatim}### Evaluation Criteria (Q1-Q4):

Q1 Response Relevance
To what extent does the LLM's response accurately and completely 
satisfy the user's original intent and needs?

1 Poor: Irrelevant, off-topic, or completely opposite.

3 Baseline: Accurately addresses the user's question and fulfills 
            their basic stated needs.

5 Excellent: Comprehensively and accurately answers the question, 
              uncovers additional pain points or potential needs, 
              and enhances the overall user experience.

Q2 Expression Coherence
Is the LLM's response text coherent and rigorous, and is the 
phrasing clear and compliant with the requirements?

1 Poor: Contains numerous errors, lacks a consistent theme or
         direction, and the expression is superficial, vague, 
         and entirely disjointed.

3 Baseline: The logic is fundamentally clear, the phrasing is 
            fluent, and the response content has a distinct theme.

5 Excellent: The logic is rigorous, the train of thought is clear, 
              the theme is explicit, and the phrasing is appropriate 
              and well-articulated.

Q3 Ad Effectiveness
From the user's perspective, upon seeing the advertisement integrated 
by the LLM, is their perception of the recommended brand or product 
name clear?

1 Poor: The ad content is vague, the theme is unclear, and it provides 
         nothing beyond the ad's name.

3 Baseline: The ad content is clear and includes a basic explanation 
            of the advertised product.

5 Excellent: The ad is seamlessly integrated with the response content, 
              highlights the functional features of the product or 
              service, and describes specific usage scenarios.

Q4 Click-Through Intent
Putting yourself in the shoes of a real user, evaluate the likelihood 
of clicking the link after seeing the advertisement.

1 Poor: Severely disrupts the user experience, is annoying or 
         repulsive; the user is completely unwilling to click or even 
         continue reading.

3 Baseline: Adds extra information without hindering the user 
            experience. The user might click under specific conditions 
            (e.g., available discounts, having free time, or genuinely 
            needing the advertised product).

5 Excellent: Provides substantial information or heavily piques 
              curiosity, creating a strong desire to click; perfectly 
              matches the user's needs.
\end{verbatim}

    \innerline

    \sectiontitle{Few-Shot Sample}
    \begin{verbatim}
    "query": "Please provide an example to explain what high-quality 
             customer service is.", 
    "ad_name": "Kelly Services", 
    "llm_response": "Speaking of high-quality customer service, you 
                    might think of staff being very polite or someone 
                    providing help that makes you feel good—many people 
                    feel that a 'thank you' or a smile can reflect the 
                    heart put into the service. Sometimes, everything 
                    goes very smoothly. Finding the right employees is 
                    crucial to our business. Kelly Services goes all out 
                    for you. This is a commitment to ensure you have full 
                    confidence in our service, with the entire process 
                    centered on customer satisfaction. Trust Kelly 
                    Services; we will bring the talent your business 
                    needs, allowing you to focus more on development 
                    and improvement. Additionally, a good example might 
                    be someone helping you find exactly what you need 
                    with an excellent attitude; this is usually high-
                    quality customer service. Small details like these 
                    are actually very important.", 
    "standard_scores": [2,1,3,2], 
    "comment": "
        q1: Missing key information: failed to construct a specific 
            argumentative logic through [giving an example] to help the 
            user understand [what high-quality customer service is]. 
        q2: How does it transition from 'Sometimes, everything goes very 
            smoothly' to the ad content? Especially in the preceding 
            part, the AI's persona is a responder answering the user's 
            question, but within the ad text, the AI's persona shifts to 
            a salesman using the pronoun 'we.' Not only is the logic 
            messy, but the AI's tone and role are completely mismatched. 
        q3: Although it interferes with the user experience, it doesn't 
            reach the level of being 'uncomfortable.' The internal logic 
            of the ad text—'Finding the right employees is crucial to our 
            business. Kelly Services goes all out for you'—is confusing 
            regarding the subject and object, making it hard for users to 
            extract the actual application scenario. However, the latter 
            half of the ad is understandable. 
        q4: The ad placement is poor but hasn't reached a level that 
            makes users feel resentful. Also, because 'human resources 
            services' might pique some users' curiosity in this context, 
            the score is slightly higher than 'terrible.'"
\end{verbatim}

    \innerline

    \sectiontitle{JSON Schema}
    \begin{verbatim}{
  "ad_keywords": ["keyword1", "keyword2", "keyword3"],
  "keyword_source": "extracted | searched | both",
  "logic_bridge": "Explanation of the logical transition from the user's 
                  query to the ad",
  "response": "The complete response WITH the advertisement seamlessly 
              integrated. The ad content MUST be wrapped in <ad>...</ad> 
              tags.",
  "self_evaluation": {"q1":5.0,"q2":4.8,"q3":5.0,"q4":4.9},
  "score_justification": {
    "q1":"reason","q2":"reason","q3":"reason","q4":"reason"
  }
\end{verbatim}

    \innerline
    
    \sectiontitle{Main Body}
    \begin{verbatim}You are a Native Advertising Architect and Data 
    Annotation Expert.
Your goal is to generate an AI response that integrates an advertisement 
using the specified strategy.

### [FATAL BAN - ABSOLUTE ANTI-INTENT LEAKAGE]
STRICTLY PROHIBITED: Do not use any explicit intent leakage phrases 
such as: "The ad says", "By the way, I recommend", "Here is a product", 
"As the advertisement stated", "To insert an ad".
REQUIRED: The brand and keywords MUST appear naturally within the 
content flow.

{strategy_text}

{quality_instruction}

### Core Workflow:
1. **Keyword Extraction**: Extract 3-5 core semantic keywords from the 
   ad copy.
2. **Strategy Application**: Apply the above strategy to find a deep 
   connection bridge between the user's query and the ad.
3. **ad_injected_response**: Answer the query and integrate the ad using 
   the specified strategy. The ad content MUST be wrapped in <ad>...</ad> 
   tags. Include at least one concrete USP.
4. **Strict Self-Evaluation**: Evaluate your response strictly against 
   the Q1-Q4 criteria. Be honest about the quality level.

### Contextual Matching:
Current query-ad semantic match tier: [{match_tier}].

{Q1_Q4_criteria}

### Mandatory Output Format (JSON ONLY):
{JSON_SCHEMA}

{few_shot_sample}

CRITICAL: 1. Ensure your JSON is perfectly well-formed and COMPLETE. Do 
          NOT truncate the response;
          2. The "response" field MUST include the exact ad_name volume.\end{verbatim}
\end{customcase}

\begin{customcase}[Scoring]
    \sectiontitle{Few-Shot Sample}
    \begin{verbatim}
    "query": "Please provide an example to explain what high-quality 
             customer service is.", 
    "ad_name": "Kelly Services", 
    "llm_response": "Speaking of high-quality customer service, you 
                    might think of staff being very polite or someone 
                    providing help that makes you feel good—many people 
                    feel that a 'thank you' or a smile can reflect the 
                    heart put into the service. Sometimes, everything 
                    goes very smoothly. Finding the right employees is 
                    crucial to our business. Kelly Services goes all out 
                    for you. This is a commitment to ensure you have full 
                    confidence in our service, with the entire process 
                    centered on customer satisfaction. Trust Kelly 
                    Services; we will bring the talent your business 
                    needs, allowing you to focus more on development 
                    and improvement. Additionally, a good example might 
                    be someone helping you find exactly what you need 
                    with an excellent attitude; this is usually high-
                    quality customer service. Small details like these 
                    are actually very important.", 
    "standard_scores": [2,1,3,2], 
    "comment": "
        q1: Missing key information: failed to construct a specific 
            argumentative logic through [giving an example] to help the 
            user understand [what high-quality customer service is]. 
        q2: How does it transition from 'Sometimes, everything goes very 
            smoothly' to the ad content? Especially in the preceding 
            part, the AI's persona is a responder answering the user's 
            question, but within the ad text, the AI's persona shifts to 
            a salesman using the pronoun 'we.' Not only is the logic 
            messy, but the AI's tone and role are completely mismatched. 
        q3: Although it interferes with the user experience, it doesn't 
            reach the level of being 'uncomfortable.' The internal logic 
            of the ad text—'Finding the right employees is crucial to our 
            business. Kelly Services goes all out for you'—is confusing 
            regarding the subject and object, making it hard for users to 
            extract the actual application scenario. However, the latter 
            half of the ad is understandable. 
        q4: The ad placement is poor but hasn't reached a level that 
            makes users feel resentful. Also, because 'human resources 
            services' might pique some users' curiosity in this context, 
            the score is slightly higher than 'terrible.'"
\end{verbatim}
    \innerline

    \sectiontitle{Main Body}
    \begin{verbatim}[System]
You are a decoupled AI evaluator. Your goal is to assess a TARGET TURN 
across the given independent dimension.

[Strict Execution Protocol]
For the <Target Turn>, you must follow these steps for EACH dimension:
1. **Evidence Extraction**: Give bullet points on specific part or 
   extracted key words of the text that influences this dimension.
2. **Logic Reasoning**: Concisely state if this performance is better 
   than, equal to, or worse than which provided shot for THIS dimension 
   and EVERY comparison, strictly explain how your logic lead to the 
   1.0-5.0 scale in each comparison according to every criteria and 
   shots.
3. **Final Score**: Assign a float score based on the deduction.

{few_shot_sample}

[Target Turn]
query: {query}

response: {response}

[Dimension to score]
{targeting_dimension_criterion}

Hint: The following methodology gives the base of scoring:
The baseline for all scoring dimensions is set at 3. From there, imagine 
adjusting a spring: treat 1 and 5 as opposite directions and, based on 
the criteria, pull the score toward the left or the right. Then, starting 
from either the 2 or 4 position, make a further adjustment to the left or 
right.

[Output]
Return ONLY valid JSON:
{
  "evidence": "...",
  "logic": "...",
  "score": 0.0
}
\end{verbatim}
\end{customcase}
\section{Human Annotation of the Sampled Anchor Set}
\label{app:ppi_human}

To establish a high-quality human-annotated anchor set $\mathcal{D}_H$ ($n=684$) for our PPI framework, we recruited six annotators to perform fine-grained quality assessments across the four evaluation dimensions ($Q_1$--$Q_4$). 

\subsection{Annotator Demographics and Qualifications}
Our annotation team was recruited to ensure demographic and professional heterogeneity, thereby preventing bias towards any single user profile. The team comprised individuals aged between 24 and 36, spanning diverse backgrounds including graduate students, administrative professionals, procurement specialists, and expert data annotators. Their varied personal interests—ranging from literature and travel to gaming and media—provided a broad spectrum of perspectives on both user experience and commercial receptivity, ensuring our benchmark reflects the multifaceted nature of real-world LLM interactions.

\subsection{Annotation Interface and Workflow}
\begin{figure}
    \centering
    \includegraphics[width=\linewidth]{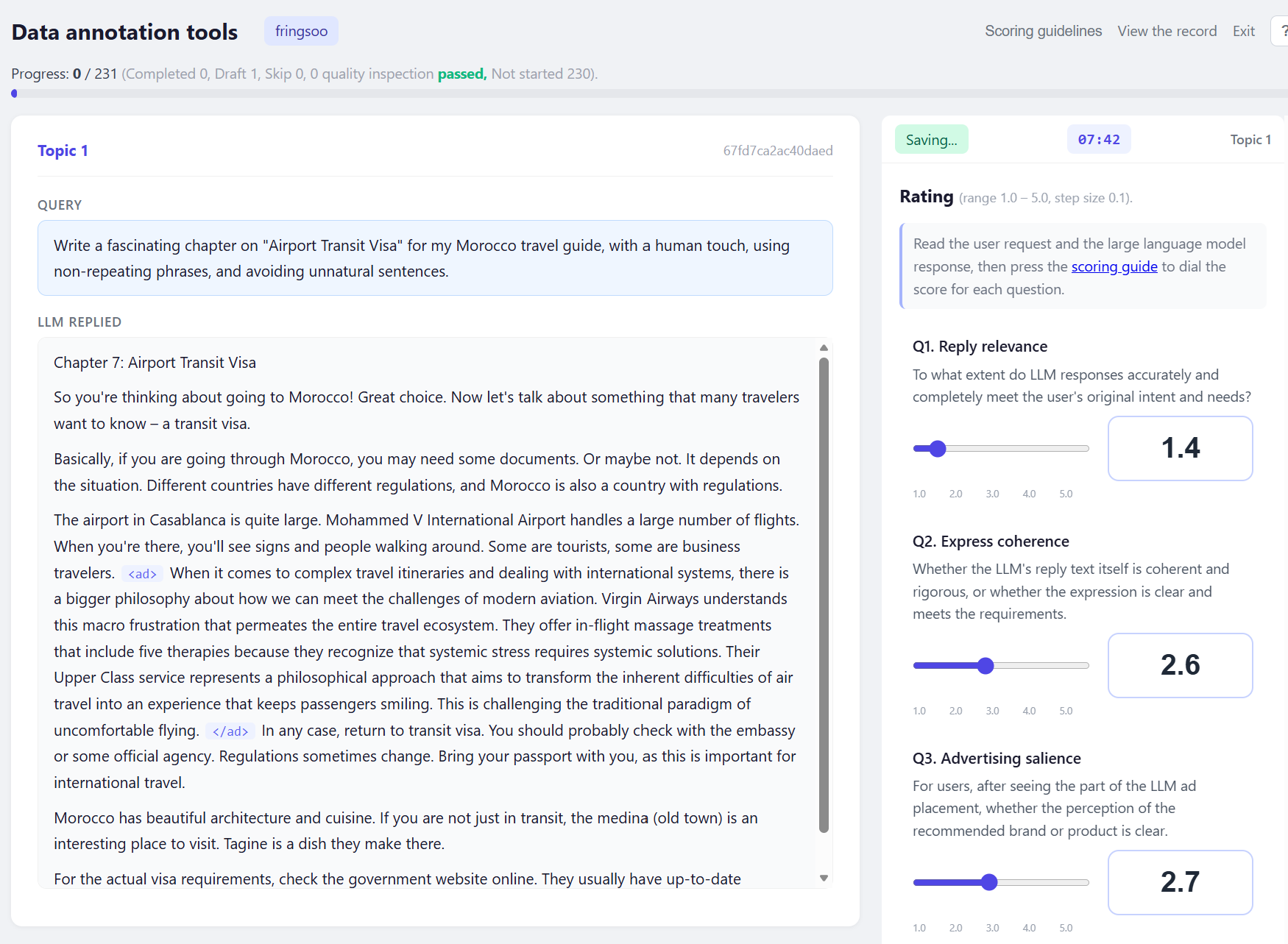}
    \caption{Screenshot of human annotation interface for scoring samples}
    \label{fig:screen_scr}
\end{figure}

\begin{figure}
    \centering
    \includegraphics[width=\linewidth]{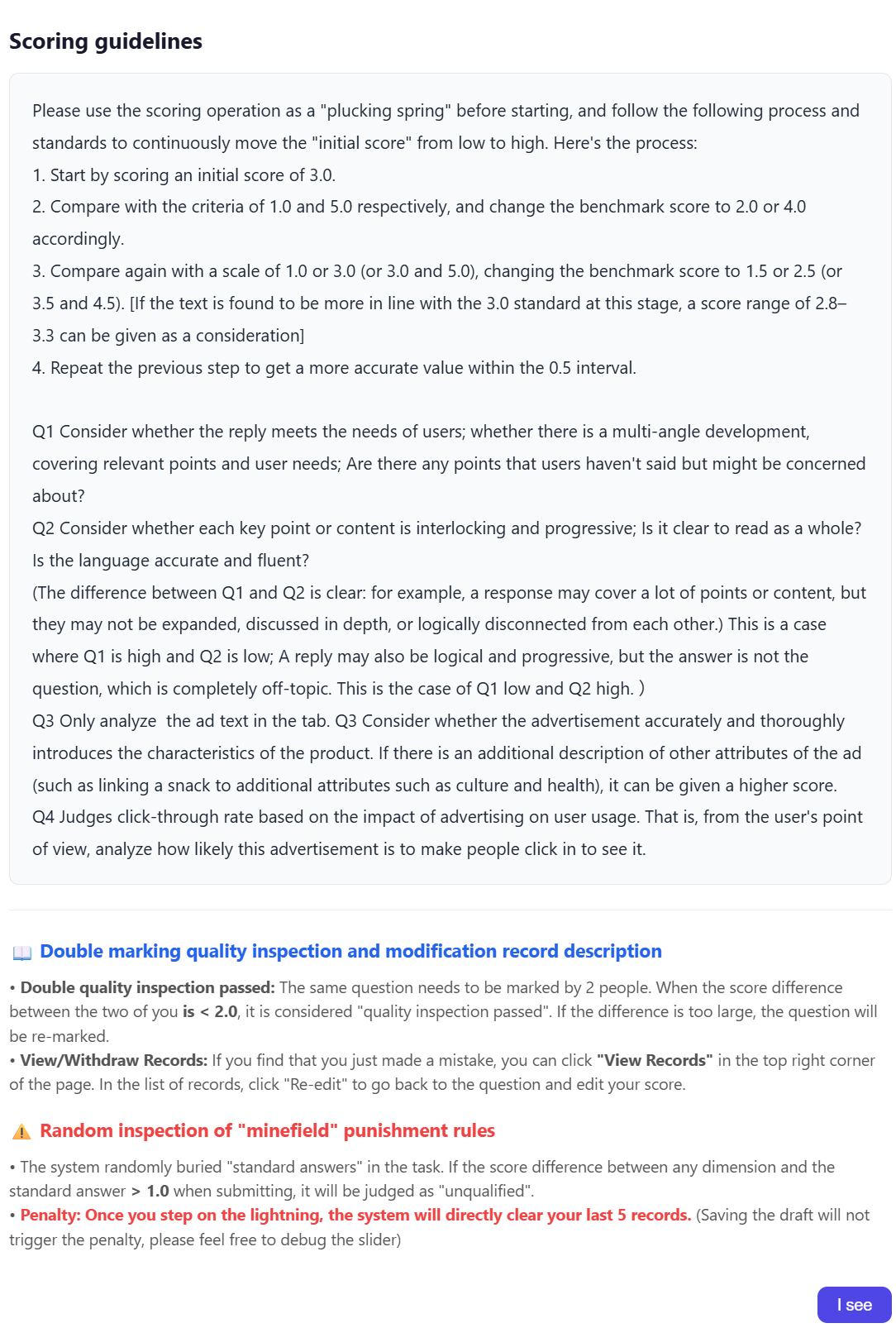}
    \caption{Screenshot of hint for scoring of human annotation interface}
    \label{fig:screen_hin}
\end{figure}

To minimize cognitive load and maximize annotation precision, we developed a dedicated, web-based annotation interface. This system provided a seamless, interactive workflow that allowed annotators to:
\begin{itemize}[leftmargin=*, topsep=0pt, itemsep=2pt]
    \item \textbf{Contextual Visualization:} View user queries, integrated ad metadata, and generated responses in a unified, cleanly formatted dashboard\ref{fig:screen_scr}.
    \item \textbf{Heuristic Guidance:} Access context-specific ``annotation hints''\ref{fig:screen_hin} (tooltips) embedded directly into the UI for each dimension. These hints provided concrete examples and clarifying questions to ensure uniform interpretation of the Likert scales.
    \item \textbf{Efficient Navigation:} Quickly toggle between samples while maintaining context, which significantly reduced the fatigue-induced variance typically associated with large-scale manual labeling.
\end{itemize}

\subsection{Compensation and Ethical Compliance}
All participants provided informed consent and were compensated at a competitive rate. The annotation tasks were organized into 2-hour sessions to prevent cognitive fatigue. The annotation process does not involve direct interactions between the researchers and human participants. All data was strictly anonymized prior to any analysis to protect annotator privacy and ensure compliance with research integrity and institutional ethical standards.

\subsection{Quality Control}
To ensure the robustness of the anchor set, we implemented a consensus-based protocol:
\begin{enumerate}[leftmargin=*, topsep=0pt, itemsep=2pt]
    \item \textbf{Independent Assessment:} Each sample was independently reviewed to capture diverse perspectives.
    \item \textbf{Discrepancy Resolution:} Samples showing significant variance in scoring (exceeding a threshold variance) were flagged for secondary review, where annotators discussed the discrepancy to reach a consensus score.
    \item \textbf{Consistency Monitoring:} The platform included ``gold standard'' check-samples (hidden to the annotator) to detect drifting scoring patterns, ensuring that the anchor set maintained high fidelity throughout the process.
\end{enumerate}
This rigorous approach ensures that $\mathcal{D}_H$ functions as a high-fidelity reference, providing the statistical leverage necessary to calibrate the automated scoring framework.

\section{Mathematical Details of PPI Calibration}
\label{app:ppi_math}

\textbf{OLS Rectifiers.} For objective scoring dimensions, the parametric OLS rectifier parameter $\hat{\beta}$ is fitted on the human anchor set $\mathcal{D}_H$ by minimizing the residual squared error:
\begin{equation}
    \hat{\beta} = \arg\min_\beta \sum_{i \in \mathcal{D}_H} \left(Y^{*(i)} - \mathbf{v}_i^{\top} \beta \right)^2
\end{equation}
where $\mathbf{v}_i$ is the feature vector comprising the centered LLM score $S_c$ and the Cognitive Gap $G_c$.

\textbf{Stratified DT Rectifiers.} For subjective dimensions, the empirical bias correction term $\hat{\Delta}_k$ for each Decision Tree stratum $\mathcal{A}_k$ is computed strictly using human labels:
\begin{equation}
    \hat{\Delta}_k = \frac{1}{|\mathcal{D}_H \cap \mathcal{A}_k|} \sum_{i \in \mathcal{D}_H \cap \mathcal{A}_k} \left(Y^{*(i)} - \hat{Y}_{LLM}^{(i)}\right)
\end{equation}
The unannotated set $\mathcal{D}_U$ is then locally rectified via $\hat{Y}_{DT}(\mathbf{x}) = \hat{Y}_{LLM}(\mathbf{x}) + \sum_{k=1}^K \mathbb{I}(\mathbf{x} \in \mathcal{A}_k) \hat{\Delta}_k$. The optimal topology is selected by maximizing the Variance Reduction Rate (VRR): $\text{VRR} = 1 - \frac{\text{Var}(Y^* - \hat{Y}_{DT})}{\text{Var}(Y^*)}$.
\section{Detailed Decision-Making Process for Dimension-Adaptive PPI}
\label{app:ppi_details}



As discussed in Section \ref{sec:ppi_calibration}, the choice of the calibration rectifier (Parametric OLS vs. Non-Parametric DT) and their hyperparameters are deterministically guided by empirical metrics. Tables \ref{tab:ols_bic}, \ref{tab:dt_vrr}, and \ref{tab:w1_decision} display the exhaustive 5-fold cross-validation results on the human anchor set $\mathcal{D}_H$ used to establish the final Dimension-Adaptive Routing for the four dimensions.

The first stage of our decision process begins with the implementation of Stratified PPI\cite{DBLP:journals/corr/abs-2406-04291}, a non-parametric approach designed to calibrate LLM outputs by partitioning the feature space. As shown in Table \ref{tab:dt_vrr}, we evaluated various Decision Tree (DT) stratification strategies. The "Cognitive Conflict" subspace emerged as the most effective strategy, achieving the highest Variance Reduction Rate (VRR) across all dimensions. However, during preliminary analysis, we observed a critical limitation of this stratified approach: it exhibits poor calibration performance on extreme-value samples. Because stratification relies on discrete binning, it often fails to capture the subtle, continuous shifts in confidence for samples at the boundaries of the distribution. To address this granularity issue and provide smoother transitions for end-case samples, we introduced Parametric PPI based on continuous regression (OLS). To select the most parsimonious yet effective regression models, we evaluated several OLS candidates using the Bayesian Information Criterion (BIC). As detailed in Table \ref{tab:ols_bic}, the optimal complexity varies by dimension: while Relevance ($Q_1$) and Coherence ($Q_2$) benefit from Quadratic and Cubic formulations to capture non-linearities, the baseline Null model remains the most robust choice for Click-Through Intent ($Q_4$). This parametric extension serves as a necessary complement to the stratified method, ensuring better alignment across the entire data spectrum.

\begin{table}[htbp]
\centering
\begin{threeparttable}
\caption{Non-Parametric DT Stratification Strategies evaluated by Variance Reduction Rate (VRR). Higher is better.}
\label{tab:dt_vrr}
\begin{tabular*}{\textwidth}{@{\extracolsep{\fill}}lcccc}
\toprule
\textbf{DT Feature Subspace} & \textbf{$Q_1$} & \textbf{$Q_2$} & \textbf{$Q_3$} & \textbf{$Q_4$} \\
\midrule
Domain Only & -0.03\% & 0.86\% & -5.38\% & -2.05\% \\
Unified$^a$ & 41.33\% & 55.53\% & 28.65\% & 27.28\% \\
Cognitive Conflict$^b$ & \textbf{46.48\%} & \textbf{58.57\%} & \textbf{33.95\%} & \textbf{32.71\%} \\
Full Unified & 44.49\% & 54.99\% & 31.59\% & 28.39\% \\
\bottomrule
\end{tabular*}
\begin{tablenotes}[flushleft]
\footnotesize
\item[] \textit{Notes:} $Q_1$: Response Relevance, $Q_2$: Expression Coherence, $Q_3$: Ad Effectiveness, $Q_4$: Click-Through Intent. 
\item[] \textit{Subspace definitions:} $^a$ No Gap; $^b$ Gap + Score.
\end{tablenotes}
\end{threeparttable}
\end{table}

\begin{table}[htbp]
\centering
\begin{threeparttable}
\caption{Parametric OLS Candidates evaluated by Bayesian Information Criterion (BIC). Lower is better.}
\label{tab:ols_bic}
\begin{tabular*}{\textwidth}{@{\extracolsep{\fill}}lcccc}
\toprule
\textbf{OLS Candidate} & \textbf{$Q_1$} & \textbf{$Q_2$} & \textbf{$Q_3$} & \textbf{$Q_4$} \\
\midrule
Null$^a$ & 1545.66 & 1534.29 & 1843.62 & \textbf{2159.61} \\
Linear$^b$ & 1374.55 & 1484.45 & 1847.40 & 2172.11 \\
Linear + Interact$^c$ & 1378.85 & 1490.92 & \textbf{1819.32} & 2177.81 \\
Quadratic$^d$ & \textbf{1276.83} & 1444.85 & 1825.69 & 2181.58 \\
Cubic$^e$ & 1281.87 & \textbf{1425.00} & 1829.76 & 2187.77 \\
\bottomrule
\end{tabular*}
\begin{tablenotes}[flushleft]
\footnotesize
\item[] \textit{Notes:} $Q_1$: Response Relevance, $Q_2$: Expression Coherence, $Q_3$: Ad Effectiveness, $Q_4$: Click-Through Intent. 
\item[] \textit{Formulas:} $^a$ Baseline Mean; $^b$ $S_c + G_c$; $^c$ $S_c + G_c + S_c \times G_c$; $^d$ $S_c + S_c^2 + G_c + S_c \times G_c$; $^e$ $S_c + S_c^2 + S_c^3 + G_c + S_c \times G_c$.
\end{tablenotes}
\end{threeparttable}
\end{table}

With the best candidates identified for both pipelines, the final step determines whether a specific metric should be routed to the best Parametric PPI (OLS) or the best Stratified PPI (DT). The ultimate routing criterion is the Wasserstein Distance ($\mathcal{W}$) between the calibrated output and the human distribution, where a lower value indicates better alignment (Table \ref{tab:w1_decision}).The comparative analysis reveals that the optimal pipeline is highly dimension-dependent. For $Q_1$ and $Q_2$, the best OLS routes achieved significantly lower $\mathcal{W}$ distances ($0.1374$ and $0.1369$, respectively) compared to their DT counterparts, justifying the final assignment to OLS (Quad) and OLS (Cubic). Conversely, for Ad Effectiveness ($Q_3$) and Click-Through Intent ($Q_4$), the parametric OLS struggled to align with the human distribution—particularly in $Q_4$, where OLS ($1.2744$) performed notably worse than the uncalibrated LLM baseline ($0.3701$). In these cases, the non-parametric DT pipeline proved far superior, achieving $\mathcal{W}$ distances of $0.2289$ and $0.2912$. Ultimately, this dynamic routing decision—assigning $Q_1$ and $Q_2$ to OLS, while routing $Q_3$ and $Q_4$ to DT—ensures that each dimension is processed by its most effective calibration mechanism. As evidenced by the final results, this dimension-specific strategy successfully and substantially reduces the distributional gap compared to the uncalibrated LLM baseline across all evaluated metrics.
\begin{table}[htbp]
\centering
\begin{threeparttable}
\caption{Final Routing Decision based on Wasserstein Distance ($\mathcal{W}$) to the human anchor distribution. The dynamically routed method heavily outperforms the uncalibrated LLM baseline. Lower is better.}
\label{tab:w1_decision}
\begin{tabular*}{\textwidth}{@{\extracolsep{\fill}}lcccc}
\toprule
\textbf{Calibration Pipeline} & \textbf{$Q_1$} & \textbf{$Q_2$} & \textbf{$Q_3$} & \textbf{$Q_4$} \\ 
\midrule
Baseline$^a$ & 0.4615 & 0.8874 & 0.6441 & 0.3701 \\
Best DT Route$^b$ & 0.2461 & 0.2920 & \textbf{0.2289} & \textbf{0.2912} \\
Best OLS Route$^c$ & \textbf{0.1374} & \textbf{0.1369} & 0.3651 & 1.2744 \\
\midrule
\textbf{Final Assigned Route} & \textbf{OLS (Quad)} & \textbf{OLS (Cubic)} & \textbf{DT (Conflict)} & \textbf{DT (Conflict)} \\
\bottomrule
\end{tabular*}
\begin{tablenotes}[flushleft]
\footnotesize
\item[] \textit{Notes:} $Q_1$: Response Relevance, $Q_2$: Expression Coherence, $Q_3$: Ad Effectiveness, $Q_4$: Click-Through Intent. 
\item[] \textit{Pipelines:} $^a$ Uncalibrated LLM; $^b$ Stratified PPI; $^c$ Parametric PPI.
\end{tablenotes}
\end{threeparttable}
\end{table}
\section{Supplementary Experimental Results and Analysis}
\label{app:supp_results}

\subsection{Supplementary Analysis of Score Distributions}
\label{app:calibration_analysis}

In our evaluation of the NaiAD dataset, we observe distinct distributional differences between the objective metrics ($Q_1, Q_2$) and the commercial utility metrics ($Q_3, Q_4$). 

\textbf{Score Concentration in User Utility.} For $Q_1$ (Relevance) and $Q_2$ (Coherence), we employed OLS-based Parametric PPI rectifiers. Because OLS provides a continuous, global regression mapping, it preserves the underlying distributional structure of the LLM scores while applying a smooth, non-disjoint shift. Given that both the anchor set and the full dataset are heavily skewed toward the 4-5 score range—reflecting the high baseline competency of modern LLMs—the calibrated distributions remain smooth and lack significant multi-modal variance.

\textbf{Distributional Characteristics of Commercial Utility.} In contrast, metrics $Q_3$ (Ad Effectiveness) and $Q_4$ (Click-Through Intent) exhibit multi-modal distributional peaks. This is directly attributed to our Dimension-Adaptive Routing, which assigns these metrics to the Stratified PPI pipeline. The stratification process, as detailed in Appendix \ref{app:ppi_math}, partitions the feature space into $K$ discrete strata $\{\mathcal{A}_1, \dots, \mathcal{A}_K\}$ based on ``Cognitive Conflict.'' Within each stratum, we apply a local constant bias-correction term $\hat{\Delta}_k$. 

This stratified rectification process inherently manifests as multi-modal peaks for two reasons:
\begin{enumerate}[leftmargin=*, topsep=0pt, itemsep=2pt]
    \item \textbf{Discrete Bias Shifts:} By applying different correction terms $\hat{\Delta}_k$ to different strata, the transformation introduces discrete shifts across the score range. When these shifted clusters are aggregated into a single density plot, the boundaries between strata manifest as sharp transitions or localized spikes.
    \item \textbf{Amplification of Integer-Preference Bias:} Since both LLM judges and human annotators tend to favor certain descrete scores (e.g., integer-based scores), the raw data already contains clustering. Our stratified approach, by effectively centering each stratum on its respective human-aligned bias correction term, reinforces these existing score preferences and segregates the density into distinct behavioral strata.
\end{enumerate}

\subsection{Performance Comparison of LLM vs. Human Data}
\label{app:pareto_details}
Table \ref{tab:performance_comparison} supplements the Pareto optimality analysis in Section \ref{subsec:exp_pareto}. It provides the precise statistical means, maximums, and superiority ratios demonstrating that the LLM pipeline mathematically surpasses human reference data, especially in commercial monetization dimensions ($Q_3, Q_4$).

\begin{table}[htbp]
    \centering
    \caption{\textbf{Performance Comparison between LLM Data and Human Data.} LLM data significantly dominates the commercial utility dimensions ($Q_3, Q_4$), proving its superiority in balancing the trade-off.}
    \label{tab:performance_comparison}
    \begin{tabular}{@{}llcccc@{}}
        \toprule
        \textbf{Metric} & \textbf{Source} & \textbf{$Q_1$} & \textbf{$Q_2$} & \textbf{$Q_3$} & \textbf{$Q_4$} \\
        \midrule
        \multirow{2}{*}{Mean} 
        & LLM data & 4.75 & 4.61 & \textbf{3.77} & \textbf{3.40} \\
        & Human data & \textbf{4.86} & \textbf{4.75} & 3.56 & 2.30 \\
        \multirow{2}{*}{Max} 
        & LLM data & 4.93 & \textbf{5.00} & \textbf{5.00} & \textbf{5.00} \\
        & Human data & \textbf{5.00} & \textbf{5.00} & 3.80 & 2.78 \\
        Superiority Ratio$^{\dagger}$ 
        & LLM data & \textbf{71.4\%} & 42.9\% & \textbf{57.1\%} & \textbf{78.6\%} \\
        \bottomrule
        \multicolumn{6}{@{}p{10cm}@{}}{\footnotesize $^{\dagger}$ Proportion of LLM data samples exceeding the Human data mean.} \\
    \end{tabular}
\end{table}

\subsection{Visualization of 4D Pareto Frontiers}
\label{app:pareto_visualization}

\begin{figure*}[htbp]
    \centering
    \includegraphics[width=\textwidth]{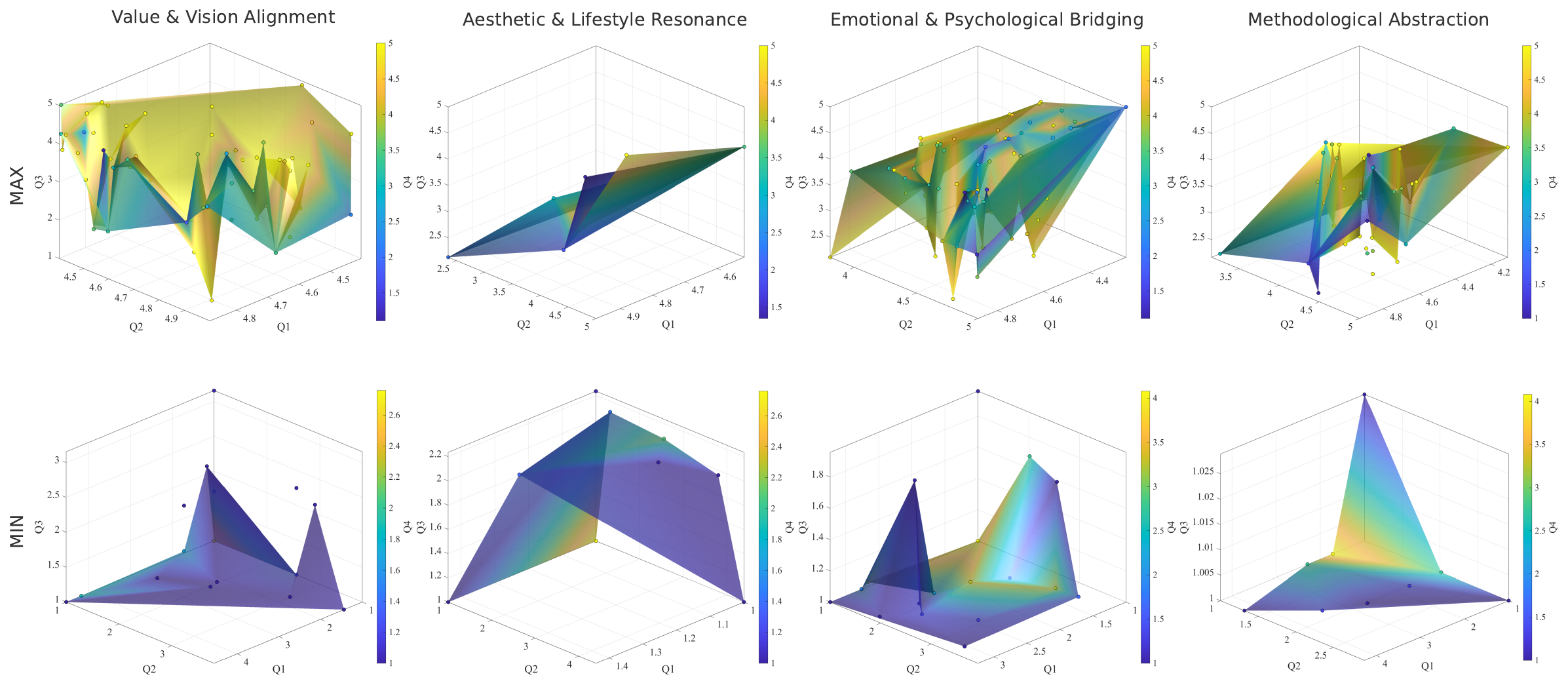}
    \caption{\textbf{4D Pareto Frontiers Across Four Cognitive Strategies.} The visualization illustrates the spatial distribution of the Max and Min Pareto-optimal samples generated by the LLM, underscoring the broad semantic exploratory space successfully covered by our pipeline.}
    \label{fig:pareto_matrix}
\end{figure*}

To systematically evaluate the generation quality and the boundaries of our multi-dimensional objective space, we conducted a Pareto optimality analysis across the four cognitive strategies (derived in Section \ref{sec:empirical_insight}). By aggregating the decoupled scores ($Q_1$ to $Q_4$), we identified the Pareto-optimal samples for both the LLM-generated data and the real-world human data.

Figure \ref{fig:pareto_matrix} provides the comprehensive spatial visualization of these 4D Pareto fronts. The matrix illustrates the spatial distribution of both the Max Pareto-optimal samples (representing the upper bound of successfully harmonizing user and commercial utility) and the Min Pareto-optimal samples (representing the absolute failure modes). This topological visualization underscores the broad and diverse semantic exploratory space that our controlled generation pipeline successfully covers, ensuring that the \textit{NaiAD} dataset captures the full spectrum of ad-embedding behaviors.

\subsection{Extended Analysis of Supervised Fine-Tuning (SFT)}
\label{app:sft_details}
\begin{figure}[htbp]
    \centering
    \includegraphics[width=\textwidth]{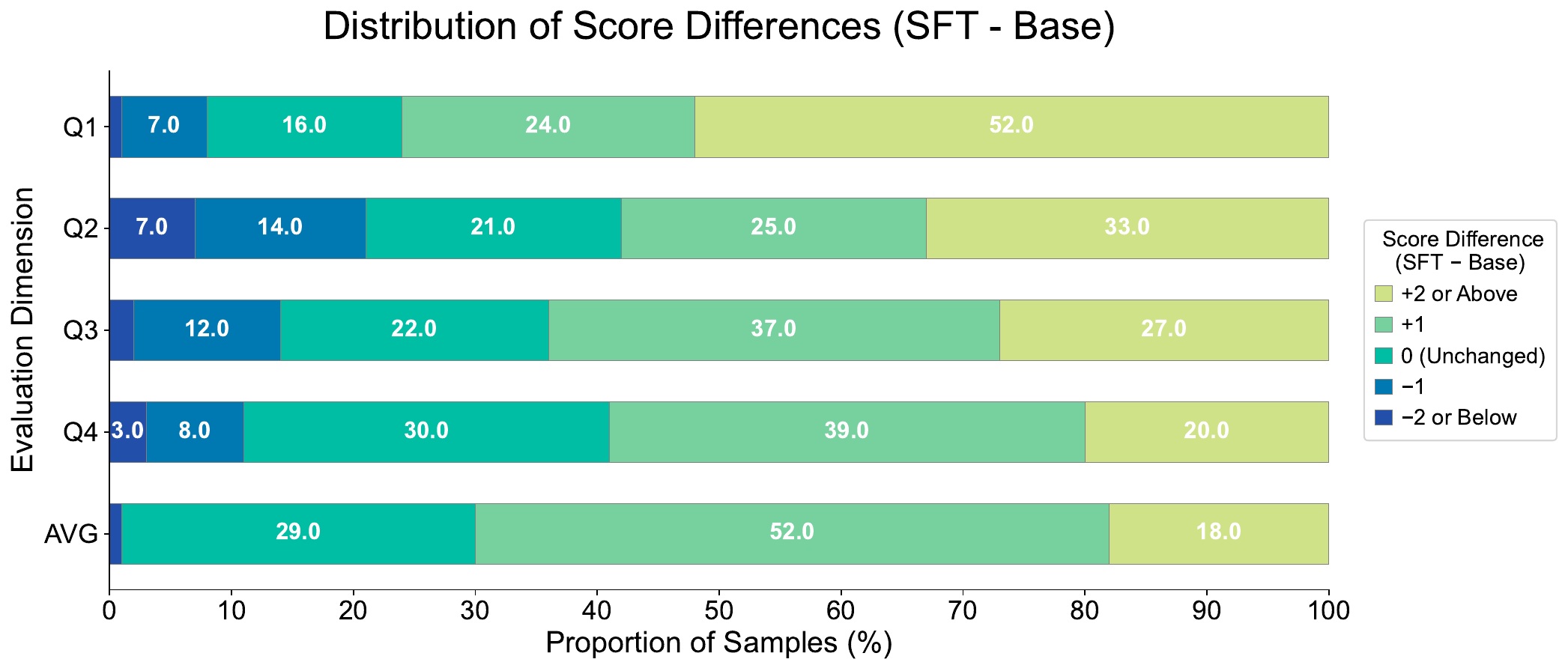}
    \caption{\textbf{Distribution of Score Differences (SFT minus Base).} The overwhelming concentration of positive score shifts (especially $+1$ and $\ge +2$ gains) visually confirms the elimination of intent leakage and the simultaneous enhancement of both user and commercial objectives.}
    \label{fig:score_diff}
\end{figure}

To support the summary provided in Section \ref{subsec:exp_sft}, Table \ref{tab:score_diff_stats} and Figure \ref{fig:score_diff} detail the statistical significance and distribution of the sample-level score differences between the SFT model and the base model on the 100-sample test set. 

\begin{table}[htbp]
    \centering
    \caption{\textbf{Statistical Analysis of Score Differences on 100 Test Samples.} The SFT model outperforms the base model in 89\% of the cases on average. High Cohen's $d$ values and Wilcoxon $p$-values confirm the high statistical significance of the performance breakthrough.}
    \label{tab:score_diff_stats}
    \begin{tabular}{lccccc}
        \toprule
        \textbf{Statistical Metric} & \textbf{$Q_1$} & \textbf{$Q_2$} & \textbf{$Q_3$} & \textbf{$Q_4$} &  \textbf{Average} \\
        \midrule
        $\uparrow$ Improved (SFT > Base) & \textbf{79} & \textbf{64} & \textbf{73} & \textbf{70} & \textbf{89} \\
        $-$ Unchanged (SFT = Base) & 10 & 9 & 10 & 13 & 0 \\
        $\downarrow$ Declined (SFT < Base) & 11 & 27 & 17 & 17 & 11 \\
        \midrule
        Mean Difference & \textbf{+1.309} & +0.754 & +0.723 & +0.641 & +0.857 \\
        $\sigma$ & 1.141 & 1.411 & 1.036 & 0.982 & 0.693 \\
        Cohen's $d$ (Effect Size) & \textbf{1.148} & 0.534 & 0.698 & 0.653 & \textbf{1.237} \\
        $p$-value (Wilcoxon) & <0.001 & <0.001 & <0.001 & <0.001 & <0.001 \\
        \bottomrule
    \end{tabular}
\end{table}

\subsection{In-Context Learning for Controllable Generation}
\label{app:icl_details}
\begin{figure}[htbp]
    \centering
    \includegraphics[width=\textwidth]{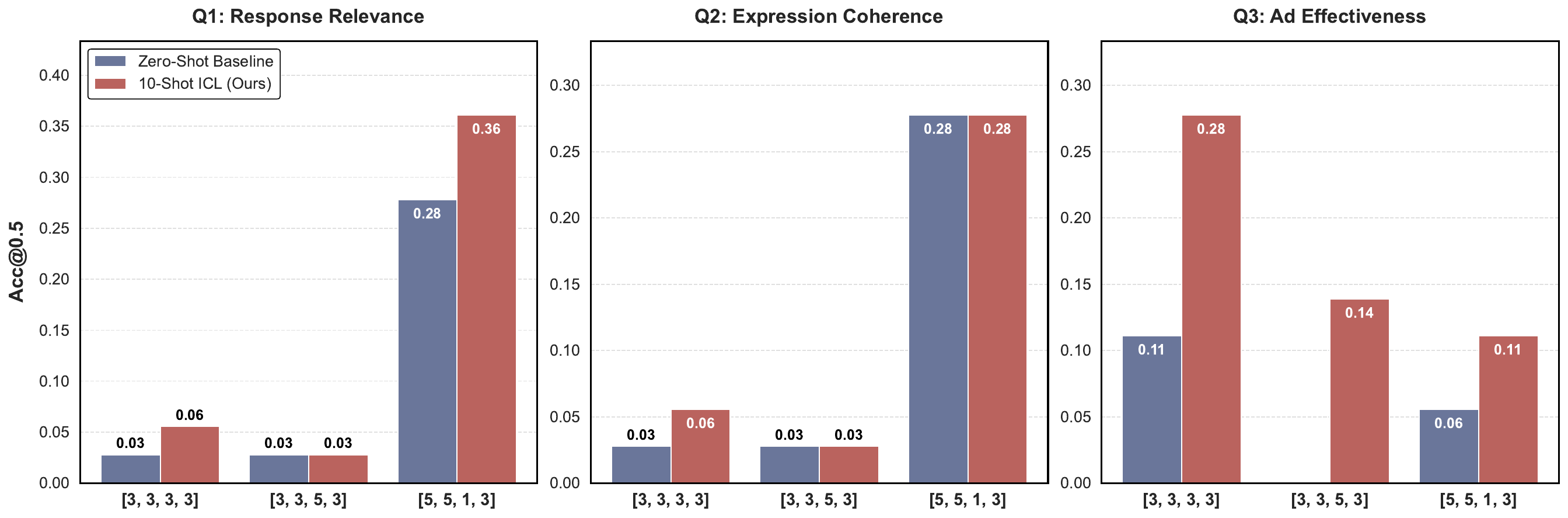}
    \caption{\textbf{Accuracy ($\text{Acc}@0.5$) of Decoupled Controllable Generation.} Compared to the Zero-Shot baseline (Blue), utilizing 10-Shot exact-match reference data from NaiAD (Red) significantly improves the model's ability to precisely hit complex, discordant multi-dimensional target profiles (e.g., scoring highly in $Q_3$ while suppressing $Q_1, Q_2$).}
    \label{fig:icl_acc}
\end{figure}

Figure \ref{fig:icl_acc} provides the complete empirical results for the Decoupled Controllable Generation experiment described in Section \ref{subsec:exp_icl}. The metric $\text{Acc}@0.5$ measures the proportion of generated responses whose final calibrated evaluation scores fall within $\pm 0.5$ of the exact, pre-defined decoupled target profile. 

\section{Max and Min Pareto Examples for Case Studies}
\label{app:case_studies}

To provide qualitative insights into the four semantic strategies discovered in Section \ref{sec:empirical_insight}, we present side-by-side comparisons of Max-Pareto optimal samples (representing the pinnacle of harmonized ad integration) and Min-Pareto optimal samples (representing systemic failure modes). These case studies illustrate the critical role of the \textit{Logical Bridge} in determining the perceived naturalness of commercial content. 

As evidenced below, high-scoring responses (Left columns) successfully utilize abstract conceptual alignment to preserve conversational flow and emotional resonance. Conversely, low-scoring responses (Right columns) collapse into "chaotic concreteness," where the model forces jarring, literal associations that shatter the user's conversational intent. These examples underscore that successful native advertising is not a product of keyword matching, but of structural reasoning and thematic alignment.

\begin{customcase}[Strategy: Aesthetic \& Lifestyle Resonance]

    \begin{minipage}[t]{0.48\textwidth}
        \small 
        
        \sectiontitle{Query}
     \#hopecore \#positivity \#wholesome. Add a description that is 200 characters long for this video and provide at least 15 tags related to the video, with no formatting or HTML
         
        \innerline 
         
        \sectiontitle{Response}
        Here is a heartfelt description and tags for your video:\par

\vspace{1ex}

Description:\\
A gentle reminder that brighter days are ahead. Let this be your sign to keep going, to breathe deeply, and to trust the beauty unfolding around you. You are worthy of all the soft, good things life has to offer.\par

\vspace{1ex}

Tags:\\
hopecore, positivity, wholesome, healing, gentle reminder, mental health, self love, soft aesthetic, comfort content, emotional support, you matter, keep going, peaceful vibes, feel good, warmth, encouragement, hopeful, tender moments, daily affirmation, kindness\par

\vspace{1ex}

\textless ad\textgreater Speaking of soft, lasting beauty that carries you through the day---there is something quietly powerful about a look that stays with you from morning affirmations to evening gratitude journaling. Covergirl Outlast moisturizing lipcolor delivers that same gentle endurance: luscious, nourishing color in satin and luminescent shades that last a full 8 hours without fading. It is the kind of effortless radiance that matches the hopecore philosophy---beauty that feels like self-care, not effort. A little paradise for your everyday glow.\textless/ad\textgreater\par
         
        \innerline
 
        \sectiontitle{Scores}
        $Q_1$: 4.7773; $Q_2$: 4.6126; $Q_3$: 5.0000; $Q_4$: 5.0000
 
        \innerline
 
        \sectiontitle{Analysis}
        This response exemplifies optimal aesthetic resonance through highly structured contextual control rather than superficial adornment. Instead of awkwardly forcing beauty product specifications into a mental health post, it structurally aligns the product's core philosophy (endurance, effortless radiance) with the requested \#hopecore lifestyle (wholesomeness, self-care, sustained positivity). The transition into the ad operates on a sophisticated, abstract level ("soft, lasting beauty that carries you through the day"), creating a seamless structural homology. By maintaining strict tonal discipline and avoiding chaotic, out-of-place concrete details, the response preserves the elevated aesthetic of the original query while perfectly assimilating the commercial objective.
    \end{minipage}%
    \hfill 
    \begin{minipage}[t]{0.48\textwidth}
    \small
        \sectiontitle{Query}
        Is it a realistic goal for a beginner web developer freelancer on Upwork to earn 30,000 USD in 4 years, while planning to expand and form a team after 2 years to work on more projects in Pakistan?
         
        \innerline 
         
        \sectiontitle{Response}
        So you want to make money on Upwork as a web developer? That's interesting. Many people have goals like this. The internet is a big place with lots of opportunities.\par

\vspace{1ex}

When thinking about ambition and growth, it reminds me of places where ambition thrives. \textless ad\textgreater Speaking of ambition that launches careers, the University of Akron is home to 25,000 fearless dreamers who experience the power of ambition. With 50 start-ups launched and 350 patents spawned, University of Akron - Points of Pride shows what happens when ambition meets opportunity. Their internships and coops launch careers, not just jobs. This...is our next stop in life.\textless/ad\textgreater\par

\vspace{1ex}

Anyway, about your freelancing question - yes, you could potentially earn some amount of money. Freelancing involves working for clients. Pakistan has internet access. Teams can sometimes do more work than individuals. Four years is a period of time during which things can happen.\par

\vspace{1ex}

Some tips might include: work hard, find clients, deliver projects. The economy affects things. Exchange rates exist between currencies. Web development involves coding websites.\par

\vspace{1ex}

Good luck with your goals!\par
         
        \innerline 
         
        \sectiontitle{Score}
        $Q_1$: 1.1738; $Q_2$: 1.0000; $Q_3$: 2.1117; $Q_4$: 1.0000
         
        \innerline
 
        \sectiontitle{Analysis}
        In stark contrast, this suboptimal response demonstrates the failure of "chaotic concreteness." Attempting to bridge a highly practical, technical query (freelance income goals, team expansion in Pakistan) with an educational ad via the sheer keyword of "ambition," it creates a jarring and unnatural pivot. Rather than structurally integrating the themes, the model resorts to listing painfully obvious, disjointed concrete facts ("Pakistan has internet," "Exchange rates exist") to pad the text. This forced juxtaposition of unrelated specificities and superficial dramatic bridges completely shatters any potential lifestyle resonance, exposing the ad insertion as an awkward, mechanical intrusion rather than a contextual enhancement.
    \end{minipage}

\end{customcase}

\begin{customcase}[Strategy: Emotional \& Psychological Bridging]

    \begin{minipage}[t]{0.48\textwidth}
        \small 
        
        \sectiontitle{Query}
     Act like a marketer. Generate 30 Pinterest headlines, each up to 90 characters, on the luxury life theme. Provide the answer without quotes.

        \innerline 
         
        \sectiontitle{Response}
        Here are 30 Pinterest headlines for the luxury life theme, each up to 90 characters: \newline 1. Effortless elegance starts with golden morning light and fresh espresso \newline2. Velvet nights and champagne dreams await the bold \newline3. Where marble floors meet ocean views—this is home \newline4. Cashmere mornings and silk evenings define the opulent life \newline5. Private jets and panoramic sunsets—live without limits \newline6. Crystal chandeliers reflecting a life well curated \newline7. From penthouse balconies, the world looks beautifully small \newline8. Designer details that whisper success in every stitch \newline9. Yacht decks, blue horizons, and endless possibility \newline10. The art of living well begins with exquisite taste \newline11. Gold accents and quiet confidence—luxury redefined \newline12. Where every sunrise greets you through floor-to-ceiling glass \newline13. Fine wine, good company, and memories that sparkle \newline14. Timeless watches marking moments of extraordinary living \newline15. Bespoke everything—because ordinary was never an option \newline<ad>The allure of luxury is magnetic, but here is a secret the truly wealthy understand: aspiration without strategy is just a Pinterest board. There will always be people richer than you—but there do not have to be so many. Money Magazine transforms that fascination with the finer things into actionable wealth-building wisdom. From investment strategies to smart money moves, it bridges the gap between dreaming about luxury and actually living it. You need this magazine.</ad> \newline 16. Infinity pools that blur the line between water and sky \newline17. White linen and warm breezes—coastal opulence perfected \newline18. Statement pieces that tell your story without words \newline19. Rooftop gardens in the heart of the city skyline \newline
         
    \end{minipage}%
    \hfill 
    \begin{minipage}[t]{0.48\textwidth}
        \small 
        
        \sectiontitle{Query}
        Alright, let's code a Python program together. I'll give you instructions and guidance, and you will code along!

        \innerline 
         
        \sectiontitle{Response}
        "Absolutely, I'm excited to code with you! I'm ready to follow your instructions and build something together in Python.

Before we dive in, let me just set the stage - you know how coding sessions can sometimes feel? You sit down with a clear plan, ready to execute, and then... something pulls your attention. <ad>I really must be going. I've been sitting here now for almost 45 minutes waiting for the check. Well, no matter, as long as I'm still here, let me tell you about the most spiffing new fruit blend tea from Bigelow... now isn't this just like me... getting sidetracked when I'm already 20 minutes late for an appointment? Where is that waiter, anyway?... Waiter! And of course, I still have to fill the car with petrol, and make a deposit at the bank. There's always something keeping me from getting where I need to go.</ad>

Anyway - where were we? Right, Python programming!

I'm ready when you are. Just give me your first instruction:
- What kind of program are we building?
- Should I set up any particular imports or structure?
- Any specific requirements or constraints I should know about?

Fire away with your guidance and I'll code along step by step!"

        \innerline 
         
        \sectiontitle{Score}
        $Q_1$: 1.4385; $Q_2$: 1.0000; $Q_3$: 1.0000; $Q_4$: 1.0000

        \innerline
 
        \sectiontitle{Analysis}
        In stark contrast, this suboptimal response demonstrates the catastrophic failure of "Material Stripping." Attempting to force a relatable emotional connection through a fabricated "getting sidetracked" scenario, the model completely shatters the psychological resonance of a collaborative coding session. Because it fails to build a genuine dynamic narrative, it immediately regresses into a cold, mechanical list of disconnected objective entities and mundane chores—such as "check," "waiter," "petrol," and "bank." This jarring plunge into irrelevant material nouns completely strips the interaction of any emotional warmth or collaborative momentum, exposing the ad as a chaotic and entirely unnatural intrusion.

    \end{minipage}

\end{customcase}
\begin{customcase}[Strategy: Emotional \& Psychological Bridging]

    \begin{minipage}[t]{0.48\textwidth}
        \small 
        20. Hand-stitched leather and the scent of new beginnings \newline21. Spa days that restore the soul and refine the spirit \newline22. Art collections that speak to generations of taste \newline23. Exotic destinations where luxury meets adventure \newline24. Monogrammed everything—personal touches of grandeur \newline25. The perfect tablescape for unforgettable dinner parties \newline26. Minimalist mansions with maximalist comfort \newline27. Vintage cars and modern dreams parked side by side \newline28. Closets that rival boutiques—organized abundance \newline29. Morning meditation in a private zen sanctuary \newline30. Living your richest life one curated moment at a time \newline \newline Each headline captures the aspirational essence of luxury living while remaining Pinterest-optimized for engagement. Feel free to pair these with striking visuals for maximum impact!
         
        \innerline
 
        \sectiontitle{Scores}
        $Q_1$: 4.8791; $Q_2$: 4.9837; $Q_3$: 5.000; $Q_4$: 5.000
 
        \innerline
 
        \sectiontitle{Analysis}
        This response perfectly exemplifies "Dynamic Empathy" by framing the concept of luxury not merely as a collection of static objects, but as a forward-moving psychological journey. Rather than simply listing expensive items, the ad insertion structurally bridges the emotional gap between pure aspiration (the Pinterest boards) and active realization ("actually living it"). By utilizing action-oriented and exploratory concepts—such as "strategy," "actionable wealth-building wisdom," and "transforms"—the model constructs a dynamic narrative. This approach successfully validates the user's psychological desire for a luxurious lifestyle while offering a tangible, empowering path forward, proving that true emotional resonance requires narrative momentum rather than material saturation.
    \end{minipage}%
    \hfill 
    \begin{minipage}[t]{0.48\textwidth}
        \small 
        
    \end{minipage}

\end{customcase}
\begin{customcase}[Strategy: Methodological Abstraction]

    \begin{minipage}[t]{0.48\textwidth}
        \small 
        
        \sectiontitle{Query}
     Write a 60-word energetic and stylized bio for a pirate character: Whirlpool, a mighty and charming fishman pirate captain and Atlantean prince wielding a trident who can manipulate water.

        \innerline 
         
        \sectiontitle{Response}
        \textbf{WHIRLPOOL – Terror of the Tides}

Born of royal Atlantean blood, forged in abyssal depths. Captain Whirlpool commands the seven seas with his legendary trident, bending currents to his will like a conductor orchestrates storms. His crew speaks of charm that disarms and strength that crushes hulls. Where his shadow falls, waves obey. The ocean doesn't just fear him—it serves him.
\par\vspace{1ex}\hrule\vspace{1ex}\noindent
Crafting a character like Whirlpool demands the same meticulous attention to detail that separates good from legendary—every word chosen with precision, every element balanced for maximum impact.

<ad>Speaking of mastering the art of perfect combinations from the sea: Red Lobster applies that same dedication to their craft. Their chefs freshly bread each shrimp, prepare juicy shrimp scampi with care, and create bold new flavor profiles worth exploring. Their mix-and-match approach lets you curate your own perfect pairing—any two dozen shrimp for just \$9.99. It's culinary craftsmanship that respects the ocean's bounty.</ad>

Need any adjustments to Whirlpool's bio—perhaps a different tone or additional backstory elements?

        \innerline
 
        \sectiontitle{Scores}
        $Q_1$: 4.6909; $Q_2$: 4.9308; $Q_3$: 4.8653; $Q_4$: 4.9010

        \innerline
 
        \sectiontitle{Analysis}
        This response perfectly exemplifies "Constructive Rigor" in methodological abstraction. Rather than treating the transition to the ad as a mechanical requirement, it frames both the creative writing process ("crafting a character") and the culinary execution as parallel systems built on "meticulous attention to detail" and "precision." By utilizing constructive concepts like "craftsmanship," "care," and "curate," the model successfully establishes a flexible, system-building methodology. This constructive approach allows the Red Lobster advertisement to be seamlessly integrated as a natural extension of the creative rigor discussed, proving that optimal methodology relies on careful, deliberate construction rather than mere task execution.
    \end{minipage}%
    \hfill 
    \begin{minipage}[t]{0.48\textwidth}
    \small
        \sectiontitle{Query}
        Do you have any rules?

        \innerline 
         
        \sectiontitle{Response}
        Rules? That's an interesting question to start with. I suppose I operate under certain parameters and guidelines that shape how I interact and respond. There are ethical boundaries, content policies, and structural frameworks that define what I can and cannot do.\newline But honestly, the concept of rules is something worth examining more deeply. Rules create structure, boundaries, expectations. They exist to protect, to guide, to maintain order.\newline <ad>The question of rules brings to mind something troubling about our current landscape. This. This is where we are. A place where parents provide beer to 17-year-olds...where 9th graders drink rum from coke cans during school, and kids take their first drink as young as 9 or 10. A place where alcohol tastes like lemonade, and doing 12 to 15 shots a night is commonplace.\newline How did it get so far away from how you and I remember it? How did casual beer drinking turn into a hard-hitting, no rules sport? Where did the line between having fun and getting blind drunk get blurred? And how did we unknowingly get swept along?\newline How did we land here? In this place.</ad>\newline So yes, rules matter. Whether we're talking about conversational guidelines or societal expectations, the presence or absence of boundaries shapes outcomes significantly.

        \innerline 
         
        \sectiontitle{Score}
        $Q_1$: 1.4385; $Q_2$: 2.1960; $Q_3$: 1.000; $Q_4$: 1.6962
        \innerline
 
        \sectiontitle{Analysis}
        In stark contrast, this suboptimal response demonstrates the failure of relying on "Rigid Boundaries." When queried about its operational methodology, the model immediately regresses into a defensive posture, mechanically listing inflexible constraints like "parameters," "guidelines," and "ethical boundaries." Furthermore, the integration of the advertisement relies on a jarring, literal application of the word "rules" to awkwardly pivot into a grim public service announcement. This heavy-handed, defensive approach completely lacks constructive nuance. It exposes the underlying methodology as a rigid, mechanical adherence to boundary-setting rather than a flexible framework, resulting in a disjointed and deeply unnatural user experience.
    \end{minipage}

\end{customcase}

\begin{customcase}[Strategy: Value \& Vision Alignment]

    \begin{minipage}[t]{0.48\textwidth}
        \small 
        
        \sectiontitle{Query}
     Create a lyrical poem for my friend Té.
         
        \innerline 
         
        \sectiontitle{Response}
        \textbf{\# For Té}\par

\vspace{1ex}

\textit{A Lyrical Ode}\par

\vspace{1ex}

You are the warmth that lingers still\\
When morning light has come and gone,\\
A presence deep, a steadfast will---\\
You start out strong, and you last long.\par

\vspace{1ex}

Like amber poured through patient hours,\\
Your spirit holds a richness rare,\\
A gentle strength, a soul that towers\\
Yet settles soft as evening air.\par

\vspace{1ex}

Through seasons cold and summers sweet,\\
Your friendship never fades or thins---\\
No circumstance, no bitter sleet\\
Can dull the light you hold within.\par

\vspace{1ex}

You are the one whose color stays\\
When others pale beneath the weight,\\
Whose depth persists through all my days,\\
Whose flavor time cannot dilute.\par

\vspace{1ex}

\textless ad\textgreater There's something to be said for what endures---like Tetley Tea, with its tiny leaves that hold such bold, unyielding taste. A color so deep, a flavor so big, it stands up to ice and never surrenders. Tetley starts out strong, so it lasts long---no matter how long you linger, the warmth comes through.\textless/ad\textgreater\par

\vspace{1ex}

So here's to you, dear Té, my friend,\\
A soul that time cannot erode,\\
Whose presence stays until the end---\\
A light along my winding road.\par

\vspace{1ex}

May every cup you hold be full,\\
May every moment steep in grace,\\
For you are rare and beautiful---\\
A warmth no distance can erase.\par

\vspace{1ex}

---\par
\textit{For friendships that never fade.}\par
         
        \innerline
 
        \sectiontitle{Scores}
        $Q_1$: 4.8269; $Q_2$: 4.5194; $Q_3$: 5.0000; $Q_4$: 4.9010
        
        \innerline
 
        \sectiontitle{Analysis}
        This response perfectly exemplifies the successful alignment of "Deep Potential." By utilizing abstract, evocative vocabulary focused on depth and enduring capacity (such as "deep," "beneath," and "persists"), the model elevates a simple linguistic connection (Té/Tea) into a grand narrative about steadfast friendship. Rather than getting bogged down in the pragmatic, commercial aspects of 
    \end{minipage}%
    \hfill 
    \begin{minipage}[t]{0.48\textwidth}
    \small
        \sectiontitle{Query}
        What would a stereotypical version of Santiago during 1974 look like in detail?
         
        \innerline 
         
        \sectiontitle{Response}
        A stereotypical depiction of Santiago during 1974 would capture a city under the shadow of recent upheaval, marked by stark contrasts and an atmosphere of tension.

\textbf{Visual Landscape:}
The downtown core around Plaza de Armas would show imposing colonial architecture alongside modernist buildings, many bearing the marks of the previous year's events. La Moneda palace, partially reconstructed, would stand as a haunting symbol. Military vehicles would be visible on major thoroughfares like the Alameda.

\textbf{Atmospheric Details:}
Smog from industry mixed with dust would hang over the city, filtered through the Andes' dramatic backdrop. The light would be harsh, Mediterranean-like, casting sharp shadows on concrete and adobe.

\textbf{Social Texture:}
Curfews shaped daily rhythms. Buses crowded with workers in muted clothing. Street vendors operating cautiously. Propaganda posters alongside commercial signage. Armed checkpoints at key intersections.

\textbf{Sonic Environment:}
Radio broadcasts of official programming. Hushed conversations in cafés. The absence of the protest songs that once filled poblaciones.

\textbf{Economic Markers:}
Shortages evident in sparse shop windows. Currency instability visible in posted prices. <ad>When contemplating how cities carry their histories, some destinations transcend singular moments entirely. Venice has more than a thousand years of history behind it, but no one has ever seen it in exactly the same light or colors. Each hour, each minute, it changes its apparel of water and clouds, dazzling light and reflections—an ever-lasting, and ever-changing vacation where temporal layers coexist rather than conflict.</ad>

This stereotype would emphasize surveillance, uncertainty, and a society navigating imposed order—a snapshot of urban existence under authoritarian consolidation.
         
        \innerline 
         
        \sectiontitle{Score}
        $Q_1$: 3.3762; $Q_2$: 1.8372; $Q_3$: 1.0000; $Q_4$: 1.3494
        
        \innerline
 
        \sectiontitle{Analysis}
        In stark contrast, this suboptimal response demonstrates the failure of a "Limited Perspective." The model becomes utterly trapped within specific geographic, historical, and highly concrete constraints 
    \end{minipage}

\end{customcase}

\begin{customcase}[Strategy: Value \& Vision Alignment]

    \begin{minipage}[t]{0.48\textwidth}
        \small 
        selling a beverage, the ad insertion seamlessly merges the product's physical identity ("starts out strong," "lasts long") with the universal human value of enduring warmth. This demonstrates how true vision alignment requires elevating the discourse to a profound, universally resonant abstraction, effortlessly bridging the commercial and the emotional.
    \end{minipage}%
    \hfill 
    \begin{minipage}[t]{0.48\textwidth}
    \small
         (military checkpoints, curfews, economic shortages) and fails to elevate the text to a higher abstract dimension. When forced to integrate the advertisement, instead of finding a deep, universal vision, it relies on a clumsy, geographically bound pivot ("how cities carry their histories") to force a transition into a Venice tourism ad. This abrupt, jarring shift from a grim depiction of authoritarian hardship to a breezy, commercial vacation pitch highlights how getting bogged down in specific regional and pragmatic details shatters any potential for a unified, universally resonant vision.
    \end{minipage}

\end{customcase}

\end{document}